\documentclass[10pt,journal]{IEEEtran}
\IEEEoverridecommandlockouts 
\usepackage{blkarray}                                      
\usepackage{algpseudocode}                                 
\usepackage{algorithm}
\usepackage{graphicx}                                      
\usepackage{amsmath}
\usepackage{amssymb}
\usepackage{amsfonts}
\usepackage{amsthm}
\usepackage[mathcal]{eucal}
\usepackage{mathrsfs}
\usepackage{booktabs}
\usepackage{enumerate}
\usepackage{multirow}
\usepackage[subrefformat=parens,farskip=0pt,justification=centering]{subfig}
\usepackage{color}
\usepackage{cite}                                          
\usepackage{comment}                                       
\usepackage{soul}                                          
\soulregister\cite7
\soulregister\ref7
\soulregister\pageref7
\usepackage{etoolbox}                                      
\usepackage{url}
\usepackage{nth}                                           
\usepackage{bm}                                            
\usepackage{courier}
\usepackage{balance}
\usepackage{threeparttable}
\usepackage{xcolor,colortbl}
\usepackage{footnote}
\usepackage{listings}
\usepackage{setspace}                                      

\usepackage{verbatim}
\usepackage[bookmarks=false]{hyperref}
\hypersetup{
    colorlinks = true,
    citecolor  = blue,
    linkcolor  = blue,
    urlcolor   = blue,
}
\usepackage{tikz}
\usetikzlibrary{patterns,snakes}
\usetikzlibrary{positioning,calc,fit,decorations.pathmorphing,shapes.geometric, shapes.gates.logic.US, calc}
\usetikzlibrary{arrows,arrows.meta,decorations.markings,shapes,shapes.arrows}
\usetikzlibrary{decorations,decorations.pathreplacing}
\usetikzlibrary{backgrounds}
\usepackage{filecontents}                                  
\usepackage{pgfplots}
\usepackage{pgfplotstable}
\usepackage{scalefnt}
\pgfplotsset{compat=newest}
\usepackage{caption}
\usepackage{cleveref}
\Crefformat{figure}{Fig.~#2#1#3}                           
\Crefname{subfigure}{Fig.}{Figs.}
\Crefname{figure}{Fig.}{Figs.}
\Crefformat{table}{TABLE~#2#1#3}                           
\usepackage[figuresright]{rotating}

\definecolor{CUHKorange}{RGB}{244,106,18} 
\definecolor{CUHKblue}{RGB}{0,111,190}    
\definecolor{CUHKgreen}{RGB}{0,127,128}   
\definecolor{CUHKred}{RGB}{228,46,36}     
\definecolor{CUHKyellow}{RGB}{198,148,34} 
\definecolor{CUHKdark}{RGB}{114,44,114}   
\definecolor{CUHKmiddle}{RGB}{144,44,144} 
\definecolor{CUHKlight}{RGB}{167,44,167} 
\definecolor{CUHKpurple}{RGB}{117,15,109}
\definecolor{CUHKgold}{RGB}{221,163,0}
\definecolor{CUHKribbon}{RGB}{244,223,176}
\definecolor{CUHKblack}{RGB}{34,24,21}

\newcommand{\tool}[1]{$\mathsf{#1}$}
\renewcommand{\bf}[1]{\textbf{#1}}
\renewcommand{\tt}[1]{\texttt{#1}}
\renewcommand{\vec}[1]{\boldsymbol{#1}}    

\newcommand{\minisection}[1]{\vspace{.1in}\noindent{\textbf{#1}}}

\usepackage{tcolorbox}
\tcbuselibrary{skins,breakable}
    {\endtcolorbox}
    {\endtcolorbox}

\usepackage[top=0.78in,bottom=0.86in,left=0.62in,right=0.62in]{geometry}
\newtheorem{myproblem}{\textbf{Problem}}
\newtheorem{mydefinition}{\textbf{Definition}}

\crefname{mytheorem}{Theorem}{Theorems}
\crefname{mylemma}{Lemma}{Lemmas}
\crefname{myclaim}{Claim}{Claims}
\crefname{myproperty}{Property}{Properties}
\crefname{mycorollary}{Corollary}{Corollaries}

\algrenewcommand\textproc{\texttt}

\makeatletter
\let\OldStatex\Statex
\renewcommand{\Statex}[1][3]{%
  \setlength\@tempdima{\algorithmicindent}%
  \OldStatex\hskip\dimexpr#1\@tempdima\relax
}
\makeatother

\RequirePackage[normalem]{ulem} 
\RequirePackage{color}\definecolor{RED}{rgb}{1,0,0}\definecolor{BLUE}{rgb}{0,0,1} 

\graphicspath{{./figs/}{../}}

\IEEEoverridecommandlockouts

\begin{document}

\title{
    \tool{DiffPattern}-Flex: Efficient Layout Pattern Generation via Discrete Diffusion
}

\author{
    Zixiao Wang,   \
    Wenqian Zhao,  \
    Yunheng Shen,  \
    Yang Bai,      \
    Guojin Chen,   \
    Farzan Farnia, \
    Bei Yu
    \thanks{
        This work is supported by The Research Grants Council of Hong Kong SAR (No.~CUHK14208021)
        and the MIND project (MINDXZ202404).
        (Corresponding author: Bei Yu)
    }
    \thanks{Zixiao Wang, Wenqian Zhao, Yang Bai, Guojin Chen, Farzan Farnia and Bei Yu are with the Department of Computer Science and Engineering, The Chinese University of Hong Kong, Hong Kong SAR.}
    \thanks{Yunheng Shen is with Tsinghua University, Beijing, China.}
}

\maketitle
\thispagestyle{plain}
\pagestyle{plain}

\begin{abstract}

    Recent advancements in layout pattern generation have been dominated by deep generative models. 
    However, relying solely on neural networks for legality guarantees raises concerns in many practical applications. 
    In this paper, we present \tool{DiffPattern}-Flex, a novel approach designed to generate reliable layout patterns efficiently. \tool{DiffPattern}-Flex incorporates a new method for generating diverse topologies using a discrete diffusion model while maintaining a lossless and compute-efficient layout representation. To ensure legal pattern generation, we employ {an} optimization-based, white-box pattern assessment process based on specific design rules. Furthermore, fast sampling and efficient legalization technologies are employed to accelerate the generation process. Experimental results across various benchmarks demonstrate that \tool{DiffPattern}-Flex significantly outperforms existing methods and excels at producing reliable layout patterns.

\end{abstract}


\section{Introduction}
\label{sec:intro}   

\IEEEPARstart{R}{eliable} very-large-scale integration (VLSI) layout pattern libraries form the backbone of various Design for Manufacturability (DFM) research, such as refining design rules \cite{he2022x,modarres1987formal,jiang2024pdrc}, optimizing Optical Proximity Correction (OPC) techniques \cite{gao2014mosaic,otto1994automated,zheng2024model}, performing lithography simulations \cite{kuang2013efficient,yu2015layout,chen2024ultra}, and detecting layout hotspots \cite{chen2019faster,yang2017layout,geng2022hotspot}. With the increasing demand for layout patterns in machine-learning-based lithography design, building a comprehensive and practical large-scale pattern library has become highly resource-intensive due to the extended logic-to-chip design cycle.

To address this challenge, a variety of rule-based and learning-based layout pattern generation methods have been introduced. Early rule-based methods \cite{reddy2018enhanced, ye2019lithoroc} augmented predefined sets of basic units through simple techniques like flipping and rotation. These units were then randomly selected and combined. However, this approach results in limited diversity and quantity of generated patterns. More recently, learning-based generative methods \cite{yang2019deepattern, zhang2020layout, wen2022layoutransformer,wang2023diffpattern,wang2024chatpattern} have demonstrated the ability to produce diverse layout patterns at a larger scale. Among these, pixel-based methods \cite{yang2019deepattern, zhang2020layout} treat pattern generation as a binary image synthesis task. While layout patterns can be expansive, the distribution of critical information within them is often sparse. To mitigate computational inefficiency, a lossless pattern representation called \textit{Squish Pattern} was introduced by \cite{gennari2014topology}, which compresses large layout patterns into smaller binary topology matrices and geometric vectors. 

In pixel-based methods, a topology matrix with continuous values is synthesized and later thresholded to produce a binary matrix. This approach introduces inefficiencies and hampers model performance. Alternatively, sequential-based methods \cite{wen2022layoutransformer} model layout patterns as polygon sequences, which are decomposed into vertices and directed edges. These methods generate new layout patterns by creating polygon sequences that are then transformed into layouts. Both pixel-based and sequential approaches rely on the generative model to learn latent regularizations from the training data to prevent generating illegal patterns that violate design rules. However, we argue that these implicit constraints learned from the data are neither flexible nor reliable. When design rules change, these models often require retraining on large-scale datasets, and even then, a significant proportion of the generated patterns may violate the rules.

In this paper, we introduce \tool{DiffPattern}-Flex, a practical and efficient pixel-based layout pattern generation framework composed of three key components: 

\minisection{Topology Generation:} Inspired by the success of diffusion models \cite{ho2020denoising,song2020denoising}, we frame the topology generation task as a denoising problem. Using a discrete diffusion model, we predict the noise that should be removed at each step. Unlike prior work that applies thresholds on continuous outputs, our method uses discrete states (e.g., $\{0, 1\}$) to represent the image tensor, naturally aligning with the discrete nature of layout patterns. This discrete output reduces overfitting and avoids the need for manual thresholding.
    
\minisection{Efficient Representation:} Building on the \textit{Squish Pattern} concept, we propose a more advanced lossless representation called \textit{Deep Squish Pattern}. This method compresses the topology matrix into a topology tensor, significantly reducing input size while expanding the channel dimension. Since diffusion models are more sensitive to input size than to the number of input channels \cite{yang2022diffusion}, this technique offers a computationally efficient solution applicable to other pixel-based generation methods.
    
\minisection{Legalization Process:} After generating topology matrices, we assign geometric vectors to them and restore legal layout patterns. To achieve this, we designed a nonlinear system that provides a legal solution for each topology matrix and is adaptable to various design rules. Our white-box legalization strategy ensures a 100\% legality rate for the generated patterns.

To further enhance diversity while maintaining pattern validity, we apply robust pattern augmentation techniques, with legality ensured via the white-box legalization process. Given the high demand for large-scale pattern libraries in downstream applications, we emphasize efficiency. To accelerate our framework, we introduce fast sampling techniques to speed up the discrete diffusion model's sampling process and employ data-guided initialization for optimal legalization. 

Our main contributions are as follows:
\begin{enumerate}
    \item We present a novel layout pattern generation method based on discrete denoising, capable of synthesizing layout topologies.
    \item We propose a lossless layout pattern representation, \textit{Deep Squish Pattern}, which improves the efficiency of pixel-based layout generation schemes.
    \item We develop a white-box legalization system that ensures all generated patterns comply with design rules.
    \item We integrate several optimization techniques that significantly speed up the generation process, achieving an $8.37\times$ improvement in sampling and a $2.48\times$ improvement in legalization.
    \item We validate our approach through extensive experiments, demonstrating state-of-the-art (SOTA) performance on benchmark datasets.
\end{enumerate}

The remainder of this paper is organized as follows. \Cref{sec:preliminaries} introduces the prior knowledge and background. \Cref{sec:DiffPattern} provides detailed explanations of our framework. \Cref{sec:acceleration} gives the details on how to enhance the diversity of generated patterns and how to accelerate both the sampling phase and the legalization phase.  \Cref{sec:experiments} presents the experimental results, followed by a conclusion in \Cref{sec:conclusion}.

\section{Preliminaries}
\label{sec:preliminaries}

\begin{figure}[tb!]
    \centering
    \includegraphics[width=0.95\linewidth]{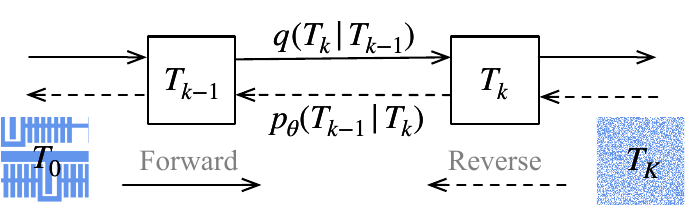}
    \caption{Illustration of denoising diffusion process.}
    \label{fig:ddpm}
\end{figure}

\subsection{Diffusion Models}
\label{sec:2.1}

Diffusion models, also known as denoising diffusion probabilistic models (DDPM) \cite{ho2020denoising,song2020denoising}, have shown great promise in generating high-quality images. These models employ a Markov chain \cite{norris1998markov} to describe both the forward and reverse diffusion processes, as depicted in \Cref{fig:ddpm}. In the forward process, a sequence of noisy samples $\vec{T}_1,...,\vec{T}_K$ is generated by progressively adding Gaussian noise to an original sample $\vec{T}_0$ over $K$ steps. The noise level is governed by a variance schedule $\left\{ \beta_k \in (0,1) \right\}_{k=1}^K$:
\begin{equation}
    \vec{q}\left(\vec{T}_k | \vec{T}_{k-1}\right) := \mathcal{N}\left(\vec{T}_k ; \sqrt{1-\beta_k} \vec{T}_{k-1}, \beta_k \vec{I}\right).
    \label{eq:ddpm-forward}
\end{equation}

When $K$ becomes large, the final noisy sample $\vec{T}_K$ approximates a Gaussian distribution, leading to the reverse diffusion process, which aims to generate new data samples from randomly drawn Gaussian noise. The reverse process aims to learn the inversion of the forward diffusion process, allowing the generation of fresh data. Since inverting this process requires an expressive model, we utilize a deep neural network with learnable parameters $\vec{\theta}$ to approximate the reverse distribution:
\begin{equation}
    \vec{p}_{\vec{\theta}} \left(\vec{T}_{k-1} | \vec{T}_k\right) := \mathcal{N}\left(\vec{T}_{k-1} ; \vec{\mu}_{\vec{\theta}} \left(\vec{T}_k, k\right), \vec{\Sigma}_{\vec{\theta}} \left(\vec{T}_k, k\right)\right),
    \label{eq:ddpm-reverse}
\end{equation}
where, $\vec{\mu}_{\vec{\theta}}$ and $\vec{\Sigma}_{\vec{\theta}}$ represent the mean vector and covariance matrix of the distribution, respectively. The subscript $\vec{\theta}$ indicates that these quantities are obtained using a neural network with trainable parameters $\vec{\theta}$.

Similar to variational autoencoders (VAEs) \cite{doersch2016tutorial}, the training objective of diffusion models is to maximize the log-likelihood function by optimizing the variational lower bound (VLB):
\begin{equation}
    L_{\mathrm{VLB}} = D_{\mathrm{KL}}\left(\vec{q}\left(\vec{T}_K | \vec{T}_0\right) \parallel \vec{p}_{\vec{\theta}}\left(\vec{T}_K\right)\right) + \sum_{k=2}^{K} L_{k} -\log \vec{p}_{\vec{\theta}}\left(\vec{T}_0 | \vec{T}_1\right),
    \label{eq:ddpm-loss}
\end{equation}
where $ L_k = D_{\mathrm{KL}}\left(\vec{q}\left(\vec{T}_{k-1} | \vec{T}_{k}, \vec{T}_0\right) \parallel \vec{p}_\theta\left(\vec{T}_{k-1} | \vec{T}_{k}\right)\right)$ and $D_{\mathrm{KL}}$ denotes the KL divergence. The term $\vec{q}\left(\vec{T}_{k-1} | \vec{T}_k, \vec{T}_0\right)$ can be computed as a Gaussian distribution derived from \Cref{eq:ddpm-forward} using Bayes' theorem.

Once the diffusion model has been trained, new samples can be generated by sampling from a standard Gaussian distribution and iteratively removing noise using the reverse process, as described by \Cref{eq:ddpm-reverse}.

\subsection{Squish Pattern Representation}
\label{subsection:quish}

\begin{figure}[tb!]
    \centering
    \includegraphics[width=0.95\linewidth]{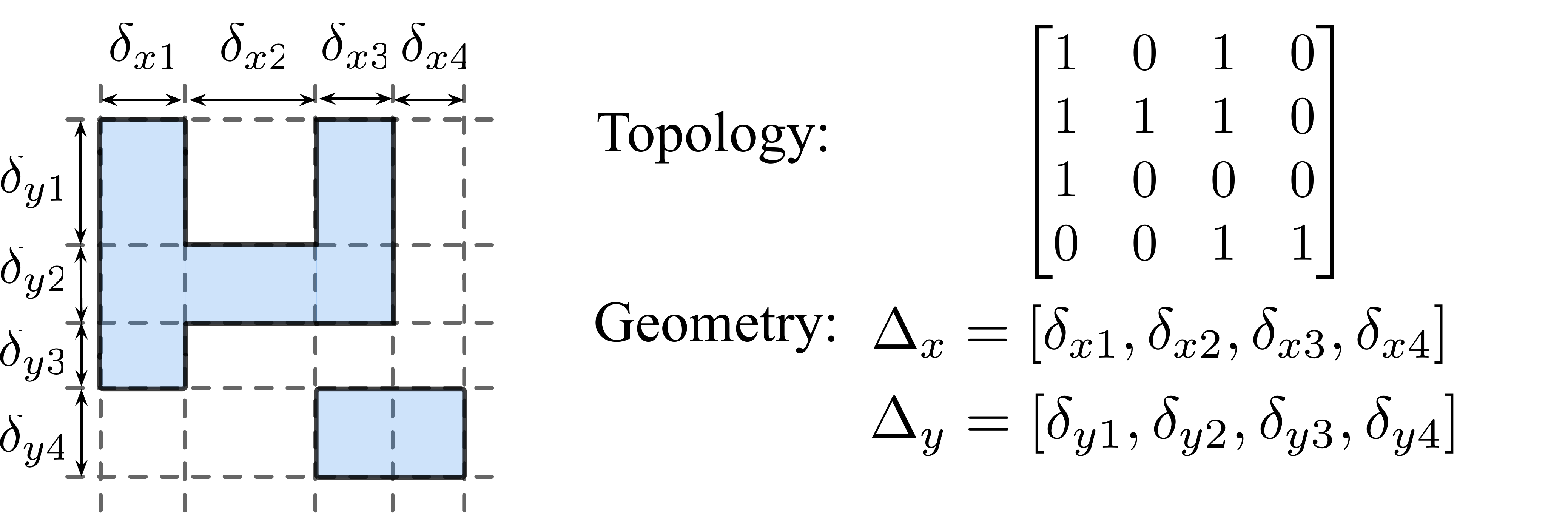}
    \caption{Squish Pattern Representation.}
    \label{fig:squishpattern}
\end{figure}

A typical layout pattern consists of a collection of polygons, which often results in information sparsity, leading to unnecessary computational complexity and potential overfitting in neural network methods. The Squish Pattern representation \cite{gennari2014topology} is an efficient, lossless encoding that compresses a layout into a topology matrix and two geometric vectors, $\Delta_x$ and $\Delta_y$, as illustrated in \Cref{fig:squishpattern}. The layout is divided into grids based on scan lines that follow the polygon edges. The interval between adjacent scan lines is recorded in the $\Delta$ vectors. Each entry of the topology matrix is binary, where one indicates the presence of a shape, and zero denotes empty space. To maintain a consistent format, the squish pattern is padded into a square shape as outlined in \cite{yang2019detecting}.

\subsection{Problem Formulation}
\label{sec:2.3}

A crucial metric for evaluating layout pattern generation is the diversity of the generated patterns. As defined in \cite{yang2019deepattern}, the complexity of a layout pattern is represented as $(c_x, c_y)$, where $c_x$ and $c_y$ are the numbers of scan lines minus one along the x-axis and y-axis, respectively. Using this definition, we can express pattern diversity as follows:

\begin{mydefinition}
The diversity of a pattern library, denoted as $H$, is defined by the Shannon Entropy of the distribution of pattern complexities:
\begin{equation}
    H = - \sum_i \sum_j P(c_{xi}, c_{yj})\log{P(c_{xi}, c_{yj})},
    \label{eq:diveristy}
\end{equation}
where $P(c_{xi}, c_{yj})$ represents the probability of a pattern with complexity $(c_{xi}, c_{yj})$ being sampled from the library.
\end{mydefinition}

Higher values of $H$ indicate greater diversity within the pattern library, signifying a broader distribution of patterns.

\begin{figure}
    \centering
    \includegraphics[width=0.95\linewidth]{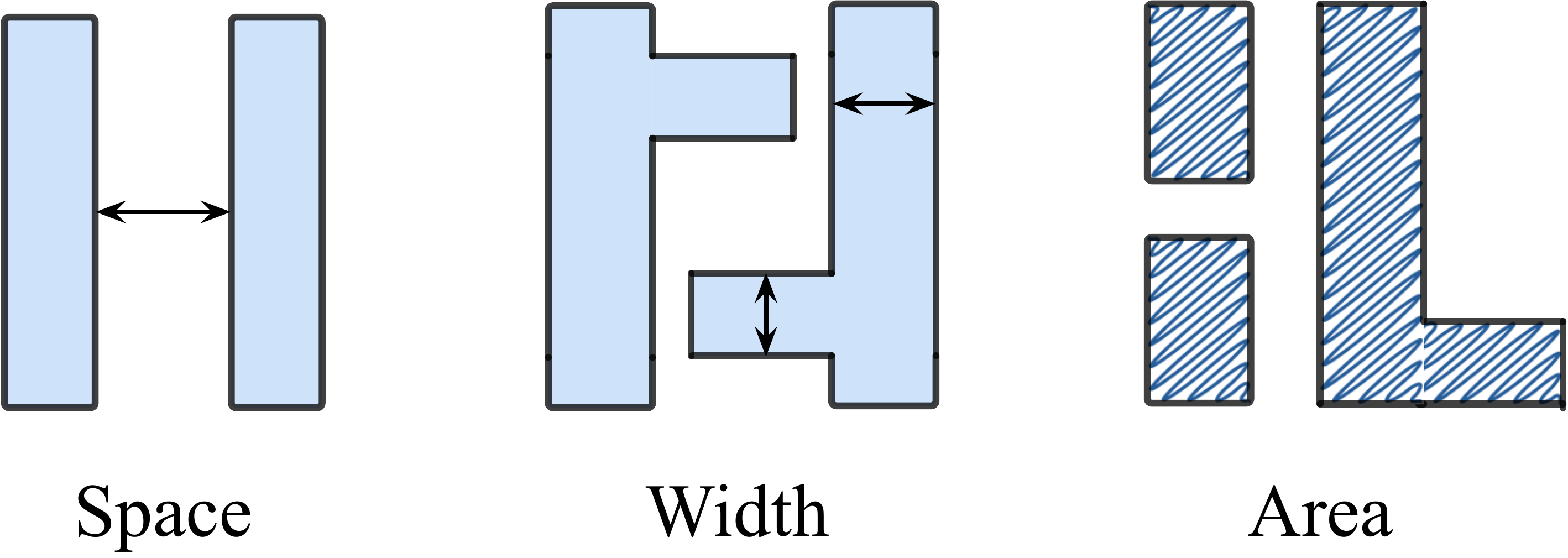}
    \caption{Illustration of design rules.}
    \label{fig:drc_rule}
\end{figure}

In addition to diversity, layout patterns must adhere to certain design rules, as outlined in \cite{zhang2020layout,wen2022layoutransformer}. As shown in \Cref{fig:drc_rule}, these design rules include `Space' (the distance between adjacent polygons), `Width' (the size of a shape in one direction), and `Area' (the area of a polygon). Based on these metrics, we define:

\begin{mydefinition}[Pattern Legality]
A layout pattern is considered \textit{legal} if it is design rule check (DRC)-clean according to the specified design rules.
\end{mydefinition}

With these evaluation metrics in mind, the pattern generation problem can be formally defined as:

\begin{myproblem}[Pattern Generation]
    Given a set of design rules and a collection of existing patterns, the objective of pattern generation is to synthesize a legal pattern library such that the diversity of the layout patterns in the library is maximized.
\end{myproblem}

\section{Reliable Discrete Pattern Generation}
\label{sec:DiffPattern}

\begin{figure*}[th!]
    \centering
    \includegraphics[width=.95\linewidth]{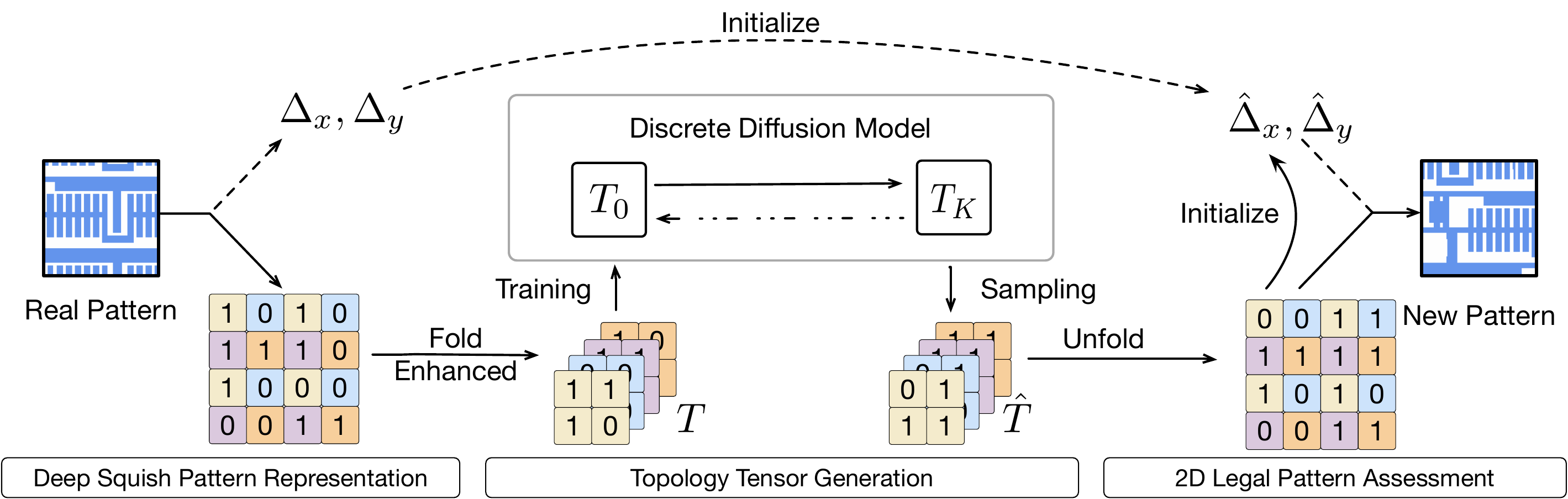}
    \caption{An illustration of the \tool{Diffpattern}-Flex framework for reliable layout pattern generation.  }
    \label{fig:pipeline}
\end{figure*}

\subsection{Overview of \tool{DiffPattern}-Flex}
As illustrated in \Cref{fig:pipeline}, our framework consists of three phases:

\minisection{Deep Squish Pattern Encoding:} For a given collection of layout patterns, we begin by deriving their deep squish pattern encoding. Each layout pattern is broken down into a topology tensor $\vec{T}$ along with two geometric vectors, $\vec{\Delta}_x$ and $\vec{\Delta}_y$.

\minisection{Generation of Topology Tensors:} The topology tensors $\vec{T}_0$ obtained from the extraction process are subsequently processed by a discrete diffusion model. In this model, noise is incrementally added to $\vec{T}$ with predetermined probabilities $q(\vec{T}_k|\vec{T}_{k-1})$, allowing the model to learn the inverse of this $K$-step noise application. To generate a new topology tensor $\hat{\vec{T}}$, a noise-infused tensor $\vec{T}_K$ is sampled, after which each component of $\vec{T}$ transitions among a set of finite states according to a predicted probability $p_\theta(\vec{T}_{k-1}|\vec{T}_k)$, eventually resulting in a plausible topology tensor $\vec{\hat{T}}_0$.

\minisection{2D Legal Pattern Evaluation:} For the generated topology tensors $\vec{\hat{T}}$, we employ an interpretable nonlinear system to allocate geometric vectors to each tensor, ensuring that the resulting layout complies with the specified \textit{Design Rules}.

\subsection{Deep Squish Pattern Representation}
\label{subsec:compact}

\begin{figure}[ht]
    \centering
    \includegraphics[width=0.95\linewidth]{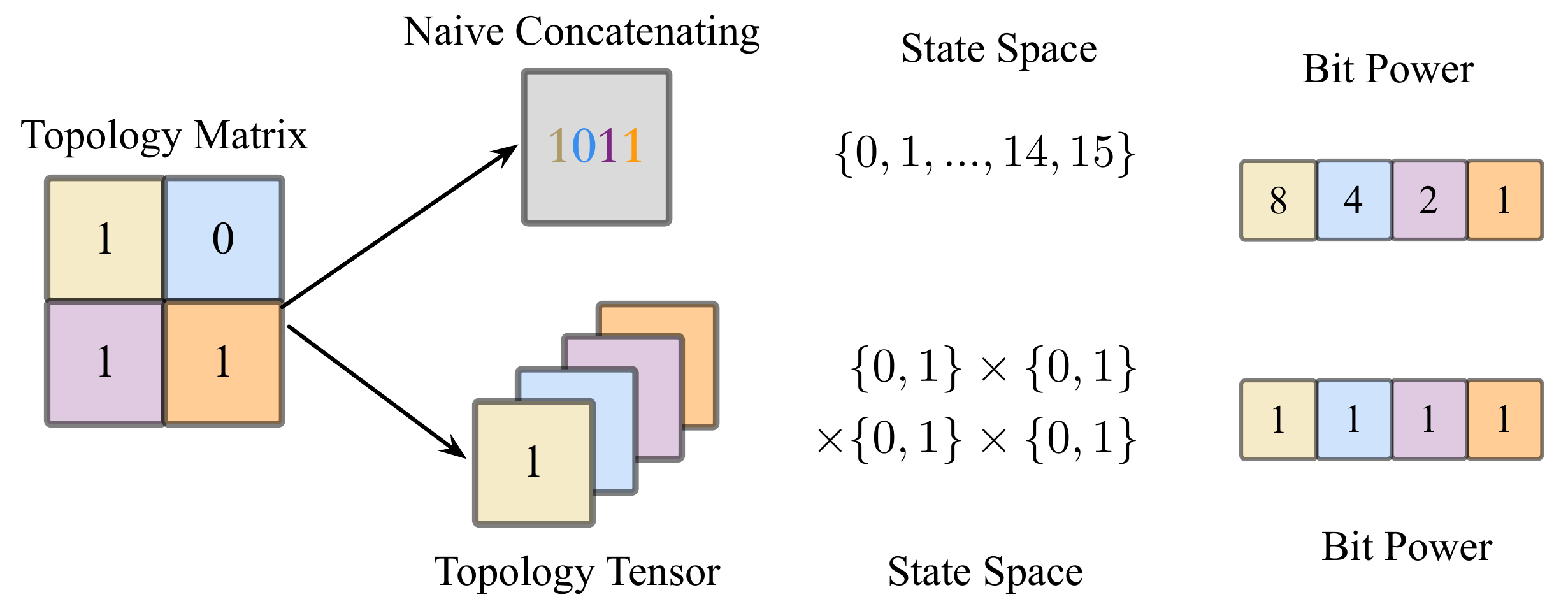}
    \caption{An illustration of the Deep Squish Pattern Encoding. The Topology Tensor provides a compact and lossless representation of the topology matrix. Simple bit concatenation leads to unequal power distribution across bits and causes exponential growth in the state space.}

    \label{fig:compact}
\end{figure}

As discussed in \Cref{subsection:quish}, the squish pattern provides a lossless representation of layout patterns, where the topology matrix is treated as a single-channel 2D binary mask, as depicted in \Cref{fig:compact} (left). However, the per-pixel information density remains suboptimal, given that the efficiency of diffusion models is more sensitive to image size than the number of pixel states. To address this issue, we propose a new representation method, the \textit{Deep Squish Pattern}, which aims to provide a more compact encoding.

Consider the example shown in \Cref{fig:compact}, where a topology matrix contains four adjacent pixels (2$\times$2), each assigned a value of either zero or one, indicating shape or space. A straightforward way to increase information density would be to encode multiple pixel bits into one. However, concatenating bits from different pixels into a single state, such as assigning values from 0 to 15 to represent all possible states for a (2$\times$2) pixel block, introduces unbalanced power to each bit. This imbalance can lead to numerical instability as the bit count increases. For instance, in a 4$\times$4 pixel case, the first bit carries a power of $2^{15}$, while the last bit holds a power of 1. Furthermore, the state space grows exponentially with the number of bits.

To resolve this, we observe that a state space of 16 ($2^4$) discrete states can be decomposed into permutations of four subspaces with two states each. Rather than assigning unequal powers to different bit positions, we assign equal weight to each bit by folding the squish topology matrix into a topology tensor $\vec{T}$ with multiple channels. In this process, a $\sqrt{C} \times \sqrt{C}$ patch from the topology matrix is transformed into a single point with $C$ channels in the tensor $\vec{T}$. The hyper-parameter $C$ is chosen to balance local information density and input size. This Deep Squish Pattern representation expands the model's effective receptive field and integrates naturally with pixel-based machine learning approaches. After the generation process is completed, the original topology matrix can be restored by unfolding the topology tensor back into its matrix form.
{
We should note that the folded topology tensor is still treated as a single sample rather than multiple separate samples during the generation process. Therefore, the deep squish pattern representation will not affect the logical resolution of the original topology matrix, preserving all information. When the topology generation is completed, the generated deep squish pattern will be flattened into a 2D topology matrix and further legalized in the subsequent steps.
}

\subsection{Topology Tensor Generation}
\label{sec:3.3}

Once we have efficiently encoded the existing layout pattern using the Deep Squish Pattern representation, the next step is to learn the distribution of existing topology tensors and generate new ones with valid topology attributes. Let ${\vec{T}_0} \in \{0,1\}^{C \times M \times M}$ represent a topology tensor extracted from existing patterns. A naive approach would treat the binary tensor as a grayscale image, learn the distribution through a diffusion model (as introduced in \Cref{sec:2.1}), and convert the generated topology to a binary one by setting a threshold, as done in previous pixel-based pattern generation methods \cite{zhang2020layout,yang2019deepattern}. However, forcing the network to learn discrete outputs (zero and one in our case) in a continuous state space from the training set is an inefficient use of the model's representational capacity. A more elegant approach is to generate discrete outputs naturally.

\minisection{Discrete Diffusion Model.}
Unlike traditional diffusion models used in computer vision, we aim to synthesize the topology of layout patterns, where each entry in the topology belongs to a discrete state. We make several key modifications to enhance the diffusion model and directly synthesize discrete topology patterns. To achieve this, we first reformulate the problem.

At the $k$-th of $K$ diffusion steps, $x_k \in \{0,1\}$ is an entry in the topology tensor $\vec{T}$. In the discrete diffusion model, a transition probability matrix $[\vec{Q}_k]_{ij} = q(x_k = j | x_{k-1} = i)$ describes the state transition probability for each $\vec{x}$ at the $k$-th diffusion step:
\begin{equation}
\vec{q}\left(\vec{x}_k \mid \vec{x}_{k-1}\right) := \operatorname{Cat}\left(\vec{x}_k ; \vec{p}=\vec{x}_{k-1} \vec{Q}_k\right),
\label{eq:d3pm-forward}
\end{equation}
where $\vec{x}_k$ is the one-hot encoded version of the entry $x_k$, $\operatorname{Cat}(\vec{x}|\vec{p})$ represents a categorical distribution over the row vector $\vec{x}$ with probabilities given by the row vector $\vec{p}$, and $\vec{x}_{k-1} \vec{Q}_k$ is the row vector-matrix product. The matrix $\vec{Q}_k$ is applied independently to each entry in the topology tensor, and $\vec{q}$ factorizes over higher dimensions as well. The Deep Squish Pattern representation is well-suited for this discrete diffusion process, as the size of the transition matrix $\vec{Q}$ increases with the number of states for each pixel. Since every entry $x$ in the topology tensor has only two states, we can efficiently model the transitions. During the reverse diffusion process, the neural network predicts the categorical distribution probability $\vec{p}_\theta(\vec{x}_{k-1}|\vec{x}_k)$ for each entry to recover the original tensor.

The choice of transition probability matrix $\vec{Q}_k$ is critical, as it should ensure that the forward process $\vec{q}(\vec{x}_k|\vec{x}_0)$ converges to a known stationary distribution as $k$ becomes large. A uniform stationary distribution is a natural choice for topology tensor generation, meaning that given any $\vec{x}_0$, the distribution of each entry $\vec{x}_k$ should follow:
\begin{equation}
    \vec{q}(\vec{x}_k|\vec{x}_0)\rightarrow \left[0.5,0.5\right],~\text{as}~k\rightarrow K.
\end{equation}
Thus, we design a doubly stochastic matrix $\vec{Q}_k$ with strictly positive entries for the topology denoising diffusion process:
\begin{equation}
\vec{Q}_k = \begin{bmatrix} 1-\beta_k & \beta_k \\ \beta_k & 1-\beta_k \end{bmatrix},
\end{equation}
where ${\beta_k} \in (0,1)$ is a hyperparameter controlling the noise level. To ensure that the model can accurately learn the original sample distribution and reach a stable distribution quickly, we adopt the classical setting from previous works \cite{ho2020denoising,austin2021structured}, using smaller noise in the early diffusion steps and larger noise in the later steps. Specifically, we use a linearly increasing schedule for $\beta_k$:
\begin{equation}
\beta_k = \frac{(k-1)\left(\beta_K-\beta_1\right)}{K-1}+\beta_1,~k = 1,...,K,
\end{equation}
where $\beta_1$ and $\beta_K$ are hyperparameters.

\minisection{Training the Diffusion Model.} To train the discrete diffusion model for topology tensor generation, the objective at each step $k$ is to minimize the following loss function:
\begin{equation}
L = D_{\mathrm{KL}}\left(\vec{q}\left(\vec{x}_{k-1} | \vec{x}_{k}, \vec{x}_0\right) \parallel \vec{p}_\theta\left(\vec{x}_{k-1} | \vec{x}_{k}\right)\right) - \lambda \log \vec{p}_{\vec{\theta}} \left(\vec{x}_0 | \vec{x}_k\right),
\label{eq:d3pm-loss-true}
\end{equation}
where $\lambda$ is a hyperparameter balancing the loss terms.

Given a topology tensor $\vec{T}_0$, we randomly sample a target step $k$ from 1 to $K$ and generate a noisy sample $\vec{T}_k$. Fortunately, we can explicitly derive that $\vec{x}_k$ follows the categorical distribution:
\begin{equation}
    \vec{q}\left(\vec{x}_k|\vec{x}_0\right) = \operatorname{Cat}\left(\vec{x}_k ; \vec{p}=\vec{x}_0 \overline{\vec{Q}}_k\right),
\end{equation}
where $\overline{\vec{Q}}_k = \vec{Q}_1 \vec{Q}_2 \ldots \vec{Q}_k$. Instead of adding noise at every step, we directly sample from this distribution to obtain $\vec{T}_k$.

After sampling $\vec{T}_k$, we feed it into the neural network along with the time step embedding $k$. The network predicts the logits of the posterior distribution $\vec{p}_{\vec{\theta}}\left(\vec{x}_0|\vec{x}_k\right)$, and $\vec{p}_{\vec{\theta}}\left(\vec{x}_{k-1} | \vec{x}_k\right)$ can be computed as follows:
\begin{equation}
\vec{p}_{\vec{\theta}}\left(\vec{x}_{k-1} | \vec{x}_k\right) = \sum_{\widetilde{\vec{x}}_0} \vec{q}\left(\vec{x}_{k-1} | \vec{x}_k , \widetilde{\vec{x}}_0\right) \vec{p}_{\vec{\theta}}\left(\widetilde{\vec{x}}_0 | \vec{x}_k\right),
\label{eq:d3pm-reverse}
\end{equation}
where $\widetilde{\vec{x}}_0$ iterates over all possible states of $\vec{x}_0$. Using Bayes' theorem and \Cref{eq:d3pm-forward}, we derive the closed form for $\vec{q}\left(\vec{x}_{k-1} | \vec{x}_k, \vec{x}_0\right)$ as:
\begin{equation}
\vec{q}\left(\vec{x}_{k-1} | \vec{x}_k, \vec{x}_0\right) = \operatorname{Cat}\left(\vec{x}_{k-1} ; \vec{p}=\frac{\vec{x}_k \vec{Q}_k^{\top} \odot \vec{x}_0 \overline{\vec{Q}}_{k-1}}{\vec{x}_0 \overline{\vec{Q}}_k \vec{x}_k^{\top}}\right),
\label{eq:d3pm-closedform}
\end{equation}
where $\odot$ is the element-wise (Hadamard) product.

At this point, all components of the loss function have been determined, allowing the diffusion model to be trained using gradient descent.

\begin{figure}[tb!]
    \centering
    \includegraphics[width=0.88\linewidth]{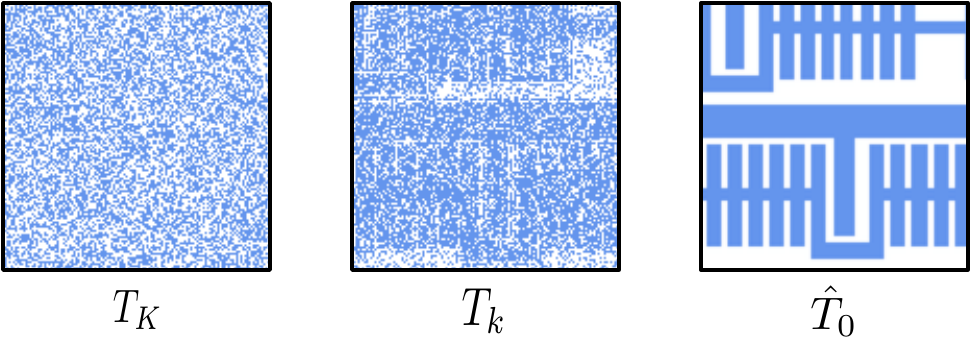}
    \caption{An illustration of the (flattened) samples from our Discrete Diffusion Model.}
    \label{fig:sampling}
\end{figure}

\minisection{Generating Deep Squish Patterns.} Once the training process is complete, we can synthesize new topology patterns by sampling a noise topology $\vec{T}_K$ from a uniform stationary distribution and iteratively removing the predicted noise using the reverse procedure. The sampling process is expressed as:
\begin{equation}
    p_\theta(\hat{\vec{T}}_0|\vec{T}_K) = p_\theta(\hat{\vec{T}}_{0}|\vec{T}_1) \prod_{k=2}^{K} p_\theta(\vec{T}_{k-1}|\vec{T}_k), \label{eq:generate}
\end{equation}
where $\vec{T}_k$ represents the estimated pattern topology at step $k$, and $\hat{\vec{T}}_0$ is the newly generated topology tensor. The denoising procedure is illustrated in \Cref{fig:sampling}. The generated topology tensor $\hat{\vec{T}}_0$ is inherently binary, with each entry being either zero or one. After sampling is complete, we flatten $\hat{\vec{T}}_0$ for legal pattern assessment.

\minisection{Topology Pre-filter.} We apply a rule-based pre-filtering process to eliminate invalid topologies, such as Bow-tie shapes, based on domain knowledge. Thanks to the high quality of the topologies generated by our discrete diffusion model, less than 0.1\% of the generated topologies are filtered out in our settings.

\subsection{2D Legal Pattern Assessment}
\label{sec:2.4}

Once the squish pattern generation is completed, the next step is to determine the legal $\Delta_x$ and $\Delta_y$ values for the generated topologies to create DRC-clean layout patterns. This decomposition of topology generation and legal pattern assessment gives \tool{DiffPattern}-Flex the flexibility to adapt to changing design rules, as discussed in \Cref{sec:4.3}. Instead of relying on black-box deep-learning methods as in previous works \cite{zhang2020layout,wen2022layoutransformer}, we employ a white-box approach to solve the problem. First, we list all constraints for each generated topology based on the design rules shown in \Cref{fig:drc_rule}, and then we formulate a nonlinear system that incorporates all these constraints, as represented in \Cref{eq:nonlinear}:
\begin{equation}
    \label{eq:nonlinear}
    \begin{cases}
        \delta_{xi}, \delta_{yj} > 0,                                                          &\forall \delta_{xi}, \delta_{yj};\\
        \sum \delta_{xi}  =\sqrt{C}M, \quad \sum \delta_{yj}  =\sqrt{C}M;                                       \\
        \sum_{i=a}^b \delta_{i} \geq\textit{Space}_\textit{min},                               &\forall (a,b)\in Set_{S};\\
        \sum_{i=a}^b \delta_{i} \geq\textit{Width}_\textit{min},                               &\forall (a,b)\in Set_W;\\
        \sum \delta_{xi}\delta_{yj}\in[\textit{Area}_\textit{min},\textit{Area}_\textit{max}], &\forall \text{Polygon};
    \end{cases}
\end{equation}
where $\textit{Space}_\textit{min}$ and $\textit{Width}_\textit{min}$ are the lower bounds for `Space' and `Width'. $\sqrt{C}M \times \sqrt{C}M$ defines the dimensions of the topology matrix, and $\textit{Area}_\textit{min}$ and $\textit{Area}_\textit{max}$ define the permissible area range for each polygon in the pattern. All constants are pattern-independent and are provided by the design rules. Both $Set_{S}$ and $Set_W$ are pattern-dependent and indicate which pairs of scan lines are constrained by design rules on `Space' and `Width', respectively.

The nonlinear system in \Cref{eq:nonlinear} can be efficiently solved using nonlinear programming algorithms or numerical methods, and typically, there are multiple possible solutions. Each solution for $\Delta_x$ and $\Delta_y$, together with the corresponding topology, constitutes a complete squish pattern representation. As discussed in \Cref{sec:4.3}, \tool{DiffPattern}-Flex can easily generate a large number of legal layout patterns from a single topology under given design rules. In rare cases, it may not be possible to find a legal solution in a limited time. Although this never occurred in our experiments (more than $1.0 \times 10^8$ attempts), we can simply discard these unsolvable cases from the generated topology set to avoid producing illegal patterns. In most cases, the choice of initial values for the nonlinear programming algorithms has minimal impact on pattern diversity and legality.

\section{Diversity and Acceleration}
\label{sec:acceleration}

\begin{figure}[t]
    \centering
    \subfloat[source]{ \includegraphics[width=0.30\linewidth]{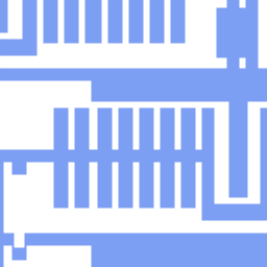} }
    \subfloat[random flip]{ \includegraphics[width=0.30\linewidth]{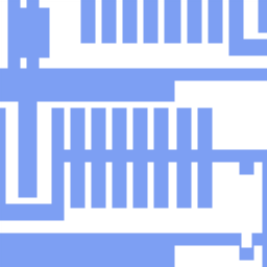} }
    \subfloat[random rotate]{ \includegraphics[width=0.30\linewidth]{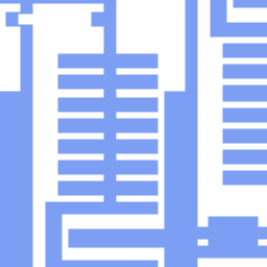} } \\
    \subfloat[symmetric mirror]{ \includegraphics[width=0.30\linewidth]{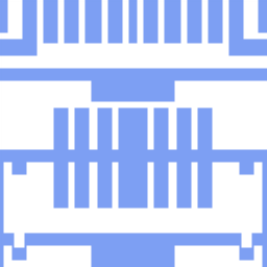} }
    \subfloat[concatenate]{ \includegraphics[width=0.30\linewidth]{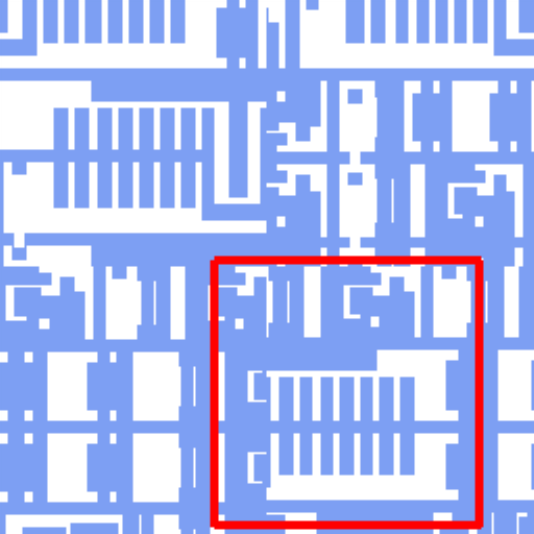} }
    \subfloat[crop]{ \includegraphics[width=0.30\linewidth]{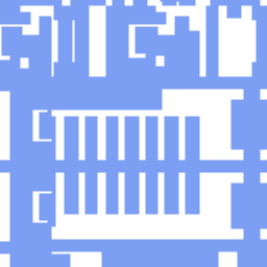} }
    \caption{Illustration of reliable pattern enhancement. In practice, multiple augmentation methods will be applied to the same pattern to enhance the data diversity. (f) denotes the random patch within the red box and is cropped from (e).}
    \label{fig:aug}
\end{figure}

\subsection{Reliable Topology Enhancement}
\label{sec:2.5}

Modern generative models are renowned for their scalability and the necessity of high-quality data. The proposed learning-based algorithm is expected to exhibit improved performance in pattern generation tasks when provided with more high-quality and diverse data. While extensive data augmentation techniques\cite{van2001art,shorten2019survey} have been introduced in recent literature, directly applying methods from general image domains to the pattern domain is not advisable. Inappropriate pattern augmentation can compromise the validity of the augmented data and potentially degrade model performance.

To ensure the reliability of enhanced topology matrices, we propose a pre-checking process using the legalization method introduced in \Cref{sec:2.4}. The reliable topology enhancement procedure is formulated as follows:
\begin{equation}
    \widetilde{\mathcal{T}} = \left\{\widetilde{\vec{T}} = \Gamma(\vec{T}) \;\middle|\; \Xi(\widetilde{\vec{T}}) = \text{Success} \right\},
\end{equation}
where $\vec{T}$ and $\widetilde{\vec{T}}$ represent the topology matrices before and after enhancement, respectively. The enhanced dataset $\widetilde{\mathcal{T}}$ is the collection of $\widetilde{\vec{T}}$. Here, $\Gamma(\cdot)$ denotes a data augmentation function, which consists of various existing enhancement techniques, and $\Xi(\cdot)$ refers to the legalization method proposed in \Cref{sec:2.4}. Although various augmentation methods may be employed, only topologies successfully legalized by the function $\Xi(\cdot)$ are used during the training of diffusion models.

In our implementation, the data augmentation function $\Gamma(\cdot)$ consists of (1) \textit{Random Flip}, (2) \textit{Random Rotation}, (3) \textit{Symmetric Mirror}, and (4) \textit{Concatenate and Crop}. Several examples of these augmentation techniques are provided in \Cref{fig:aug}. These methods were chosen because they do not introduce excessive perturbations, such as meaningless random noise, into the topology matrices. Overly noisy data is rare in practical scenarios and can negatively impact downstream tasks\cite{natarajan2013learning,wang2023truncate}.

Pre-checking with the legalization function $\Xi(\cdot)$ is crucial due to the inevitable noise introduced by augmentation methods. A potential concern is the increased time required for optimizing \Cref{eq:nonlinear}, which scales with the size of the topology matrix $\vec{T}$. This issue will be further discussed in the next subsection.

\subsection{Acceleration}
\label{sec:2.6}

Given the large demands of pattern libraries, the efficiency of the pattern generation framework is crucial. In \tool{DiffPattern}-Flex, both the topology-generation phase and the legalization phase can be further accelerated as detailed below.

\begin{figure}[t]
    \centering
    \includegraphics[width=0.95\linewidth]{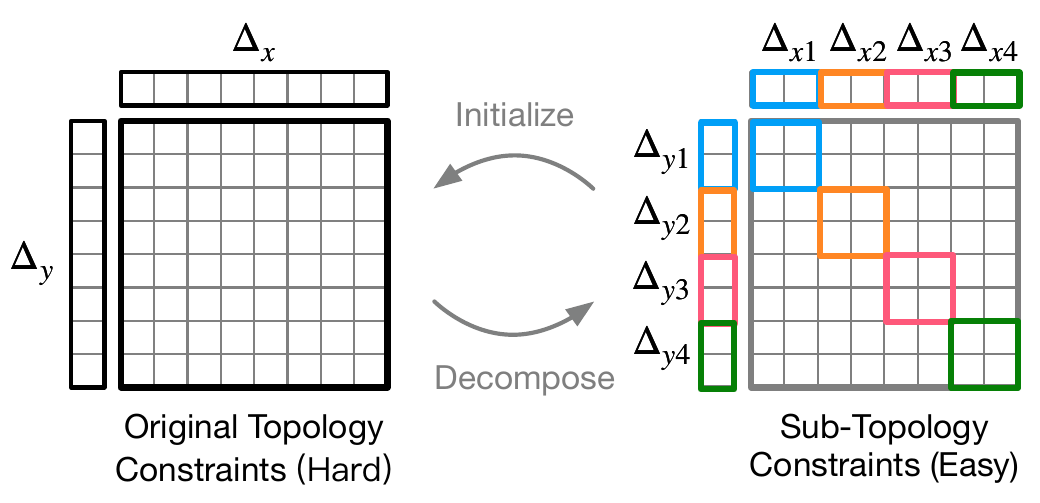}
    \caption{Illustration of legalization acceleration. Each diagonal block denotes an independent sub-problem, and all sub-problems can be solved in parallel.}
    \label{fig:block}
\end{figure}

\minisection{Fast-Sampling.} When generating new topologies from well-trained models, a straightforward approach is to gradually reduce the noise in the topology $\vec{T_k}$ and reduce the step $k$ by one during inference, as explained in \Cref{eq:d3pm-reverse} and \Cref{eq:generate}. However, considering the large step number used in practice ($K=1000$ in our implementation), the generation process can be time-consuming. To accelerate the inference procedure, we observe that it is possible to perform inference with $m$ steps at a time by modifying \Cref{eq:d3pm-reverse} into
\begin{equation}
\vec{p}_{\vec{\theta}}\left(\vec{x}_{k-m} | \vec{x}_k\right) = \sum_{\widetilde{\vec{x}}_0} \vec{q}\left(\vec{x}_{k-m} | \vec{x}_k , \widetilde{\vec{x}}_0\right) \vec{p}_{\vec{\theta}}\left(\widetilde{\vec{x}}_0 | \vec{x}_k\right),
\label{eq:d3pm-reverse-mstep}
\end{equation}
where $\vec{q}\left(\vec{x}_{k-m} | \vec{x}_k, \widetilde{\vec{x}}_0\right)$ has a closed form by extending \Cref{eq:d3pm-closedform} into
\begin{equation}
\vec{q}\left(\vec{x}_{k-m} | \vec{x}_k, \vec{x}_0\right) = \operatorname{Cat}\left(\vec{x}_{k-m}; \vec{p} = \frac{\vec{x}_k \vec{Q}_k^{\top} \odot \vec{x}_0 \overline{\vec{Q}}_{k-m}}{\vec{x}_0 \overline{\vec{Q}}_k \vec{x}_k^{\top}}\right).
\label{eq:d3pm-closedform-mstep}
\end{equation}

\begin{table*}[tb!]
    \centering
    \caption{Numerical results on pattern diversity and legality. `Real Patterns' are from the ICCAD 2014 contest. `-' denotes an inapplicable result.
    {Results of learning-based baseline are from previous works \cite{wang2023diffpattern}.}}
    \label{tab:diversity}
    {
        \begin{tabular}{c|c|cc|cc}
            \toprule
            \multirow{2}*{Set/Method}&\multirow{2}*{Generated Topology}&\multicolumn{2}{c|}{Generated Patterns}&\multicolumn{2}{c}{Legal Patterns}\\
            &&Patterns & Diversity ($\uparrow$)  &Legality ($\uparrow$) & Diversity ($\uparrow$)\\ \midrule
            \rowcolor{green!10} Real Patterns&-&-&-&13869&10.777\\            
            CAE\cite{yang2019deepattern}&100000&100000&4.5875&19&3.7871 \\
            VCAE\cite{zhang2020layout}&100000&100000&10.9311&2126&9.9775 \\
            CAE+LegalGAN\cite{zhang2020layout}&100000&100000&5.8465&3740&5.8142 \\
            
            {Rule-based\cite{ye2019lithoroc}}& {-} &{100000}&{10.256}&{24474}&{10.066}\\
            VCAE+LegalGAN\cite{zhang2020layout}&100000&100000&9.8692&84510&9.8669 \\
            LayouTransformer\cite{wen2022layoutransformer}&-&100000&10.532&89726&10.527 \\
            \tool{DiffPattern} \cite{wang2023diffpattern}&100000&100000&10.815&\textbf{100000}&10.815 \\
            Ours&100000&100000&\textbf{11.713}&\textbf{100000}&\textbf{11.713}\\
            \bottomrule 
        \end{tabular}
    }
\end{table*}

By performing inference with $m$ steps at a time, the topology generation can be roughly accelerated by a factor of $m$. However, the choice of $m$ is a trade-off between the quality of topologies and speed. A higher $m$ accelerates inference at the cost of reduced quality, while a lower $m$ provides better quality at the expense of slower inference.

\minisection{Efficient Implementation of Legalization.} The time required for the legalization phase increases significantly with the size of the topology matrix \cite{kraft1988software}. Several works \cite{lawrence2001computationally, boggs1995sequential,modarres1987formal} have explored acceleration methods for this process in optimization literature. Considering the special form of constraints in our case, some customized techniques can be applied to accelerate the optimization process.

Empirically, we found that a good initialization improves the convergence of the optimization process. In practice, when the generated patterns follow the same design rules as the source patterns, we randomly select a pair of existing geometric vectors from the dataset as the starting point. This empirically accelerates the convergence of the nonlinear programming algorithm. A potential explanation is that existing geometric vectors are more likely to partially satisfy the constraints of the optimization problem. However, when design rules change or suitable geometric vector pairs are unavailable, a general initialization method becomes critical.

To address this issue, we extend the idea of `partially satisfying constraints' and initialize the optimization problem using solutions from several independent sub-problems, as illustrated in \Cref{fig:block}. Specifically, we solve several independent sub-problems, each constrained by one diagonal block in the original topology matrix. The solving processes of sub-problems are independent and can be efficiently optimized in a parallel fashion. The solution of each sub-problem satisfies the constraints within its block and is concatenated with others to initialize the original optimization problem, thus accelerating the process. 

\begin{figure*}[tb!]
    \centering
    \subfloat{ \includegraphics[width=0.15\linewidth]{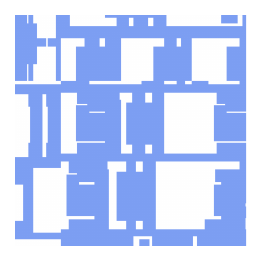} }
    \subfloat{ \includegraphics[width=0.15\linewidth]{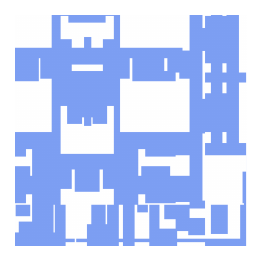} }
    \subfloat{ \includegraphics[width=0.15\linewidth]{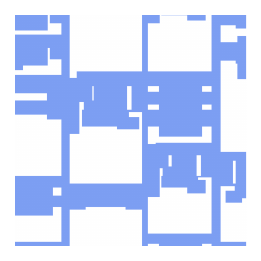} }
    \subfloat{ \includegraphics[width=0.15\linewidth]{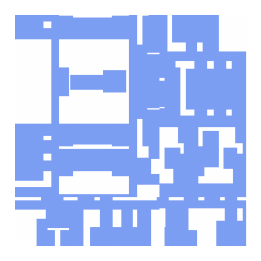} }
    \subfloat{ \includegraphics[width=0.15\linewidth]{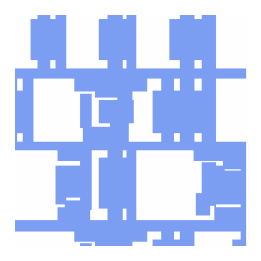} }
    \subfloat{ \includegraphics[width=0.15\linewidth]{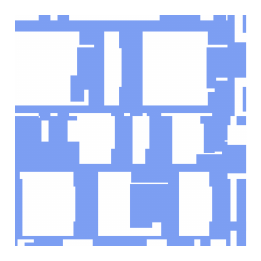} } \\
    \subfloat{ \includegraphics[width=0.15\linewidth]{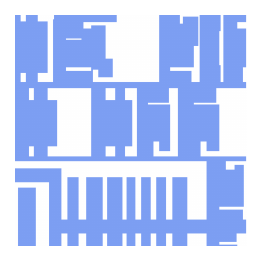} }
    \subfloat{ \includegraphics[width=0.15\linewidth]{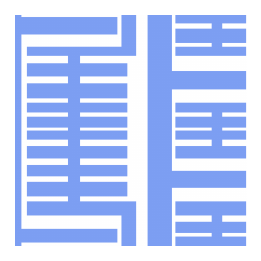} }
    \subfloat{ \includegraphics[width=0.15\linewidth]{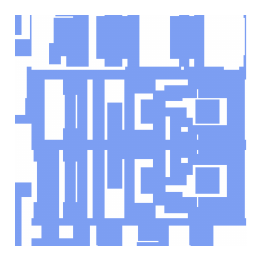} }
    \subfloat{ \includegraphics[width=0.15\linewidth]{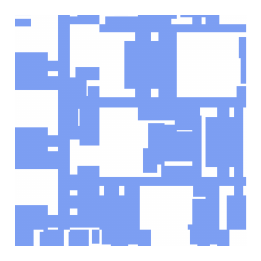} }
    \subfloat{ \includegraphics[width=0.15\linewidth]{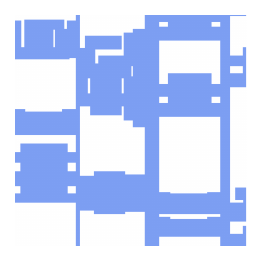} }
    \subfloat{ \includegraphics[width=0.15\linewidth]{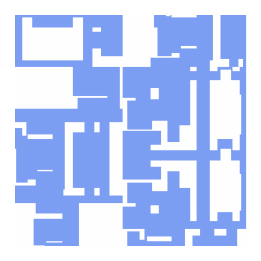} } \\
    \subfloat{ \includegraphics[width=0.15\linewidth]{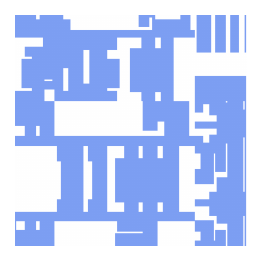} }
    \subfloat{ \includegraphics[width=0.15\linewidth]{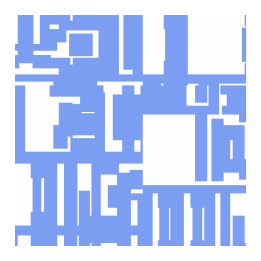} }
    \subfloat{ \includegraphics[width=0.15\linewidth]{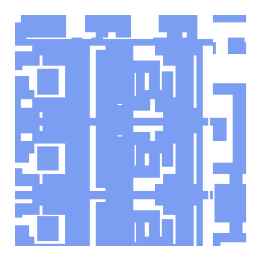} }
    \subfloat{ \includegraphics[width=0.15\linewidth]{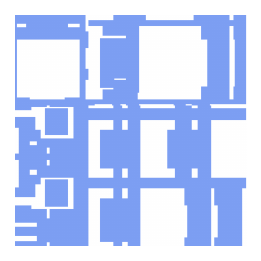} }
    \subfloat{ \includegraphics[width=0.15\linewidth]{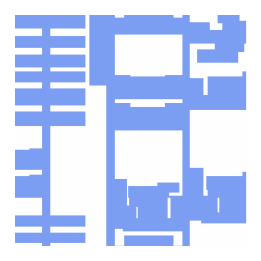} }
    \subfloat{ \includegraphics[width=0.15\linewidth]{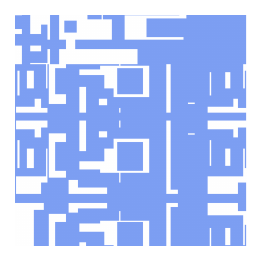} } \\

    \caption{Some randomly selected examples generated by our method. }
    \label{fig:generated}
\end{figure*}

\section{Experimental Results}
\label{sec:experiments}

\subsection{Experimental Setup}

\minisection{Datasets.} In line with prior research \cite{zhang2020layout, wen2022layoutransformer}, we utilize a dataset comprising small layout pattern images sized $2048\times2048$ $nm^2$, derived by segmenting a $400\times160$ $\mu m^2$ layout map from the ICCAD 2014 contest. The extracted topology tensor is maintained at a fixed size of 16$\times$32$\times$32 with $C=16$, as used in the Deep Squish Pattern Representation framework.

\begin{figure}[tb!]
    \centering
    \subfloat[]{ \includegraphics[width=0.30\linewidth]{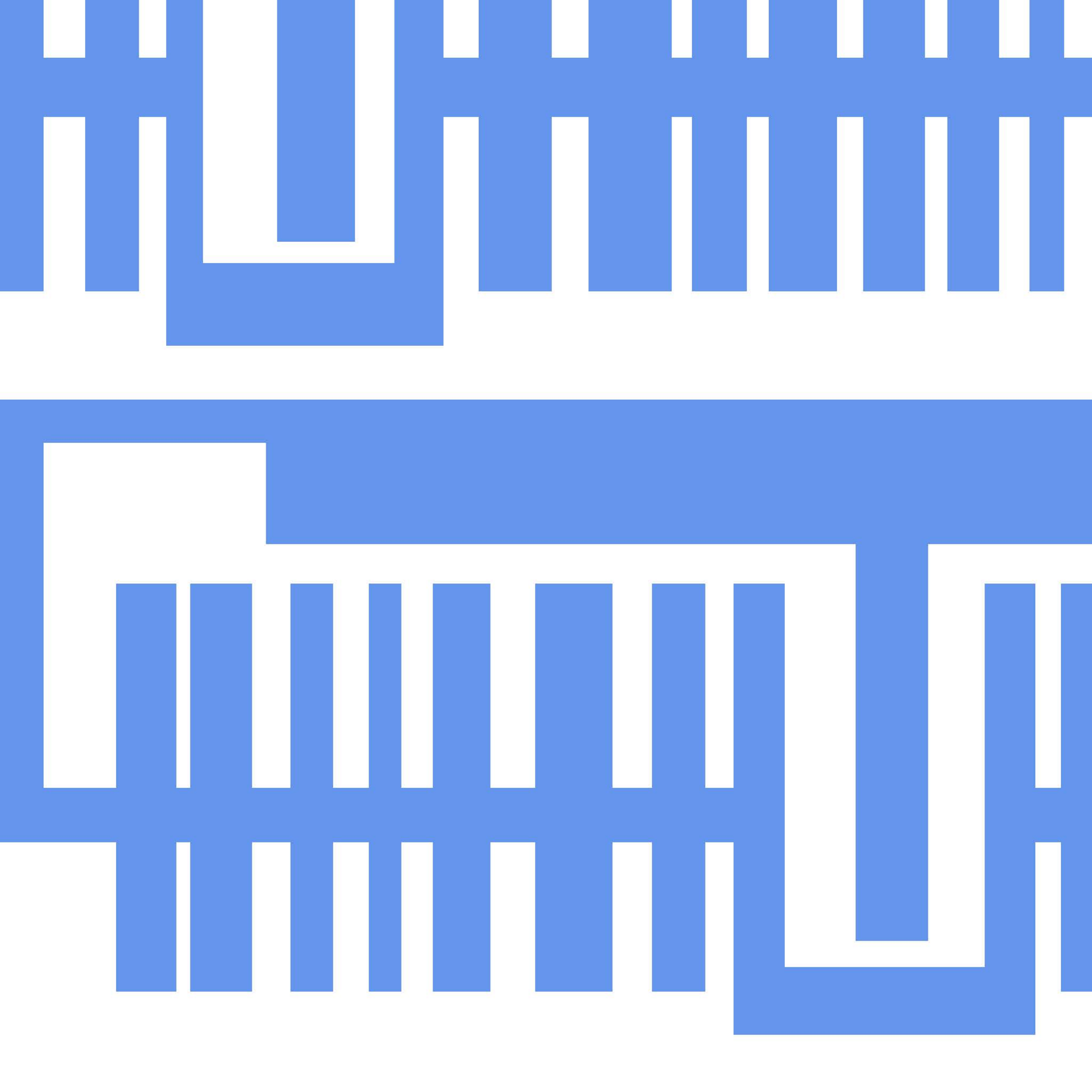} }
    \subfloat[]{ \includegraphics[width=0.30\linewidth]{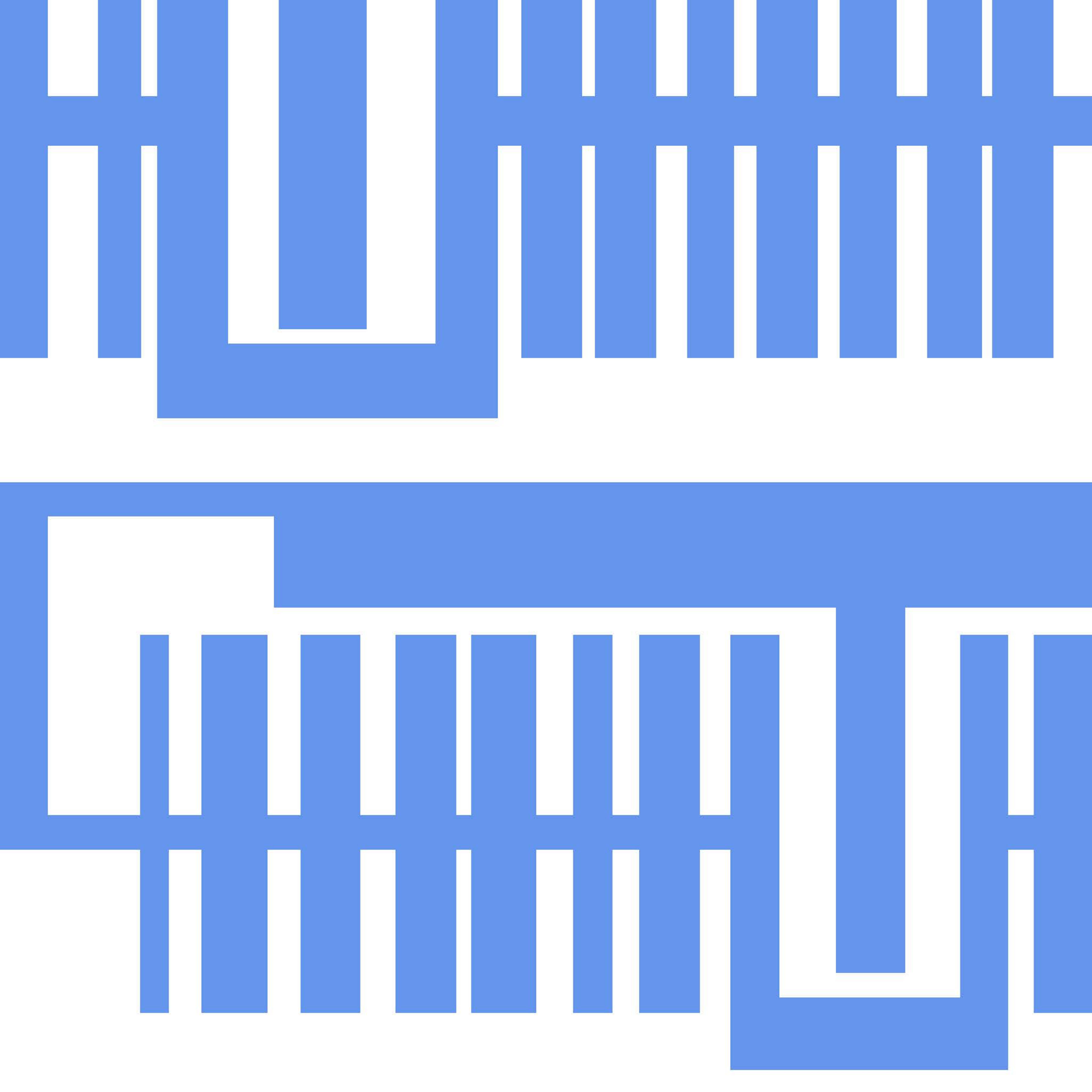} } 
    \subfloat[]{ \includegraphics[width=0.30\linewidth]{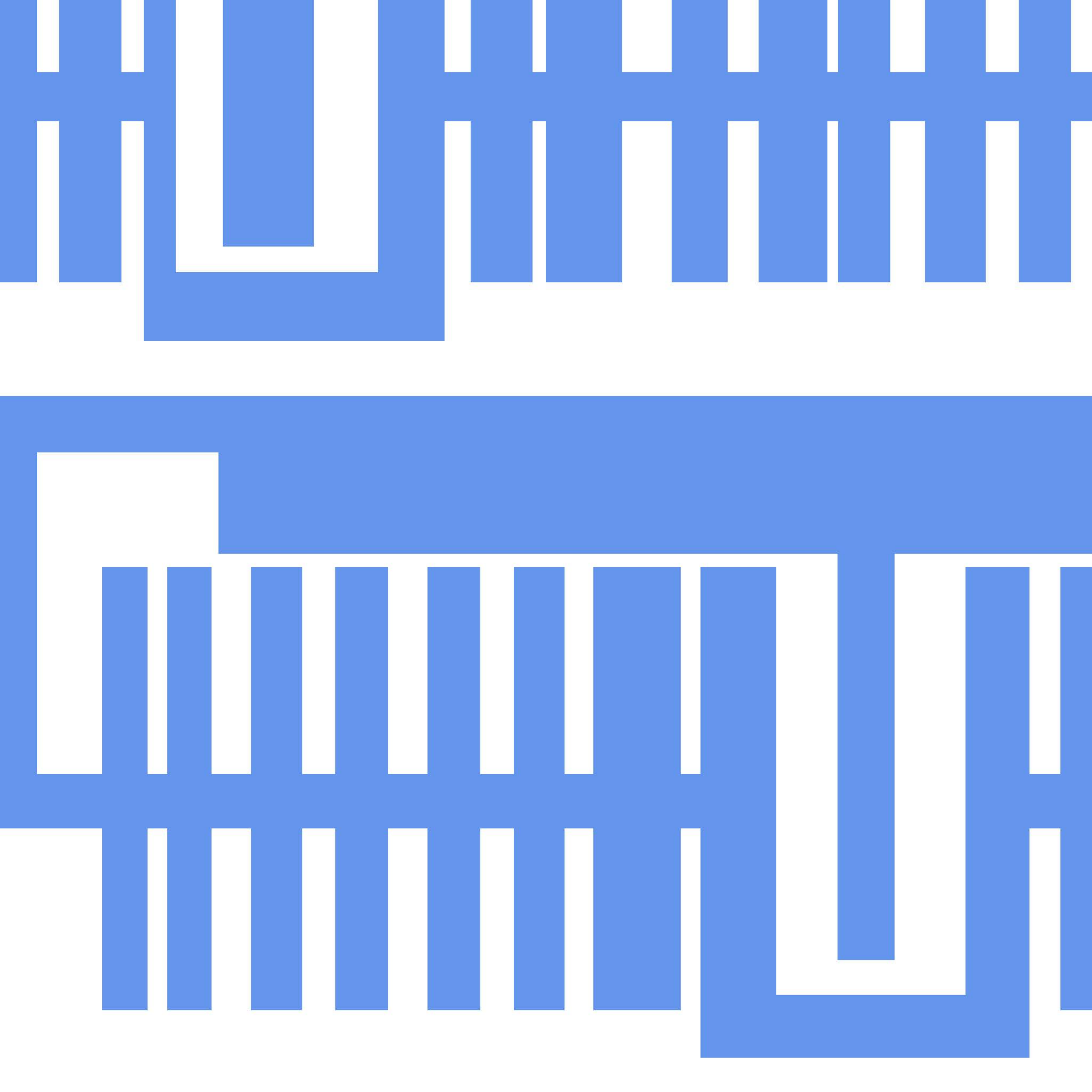} } \\
    \subfloat[]{ \includegraphics[width=0.30\linewidth]{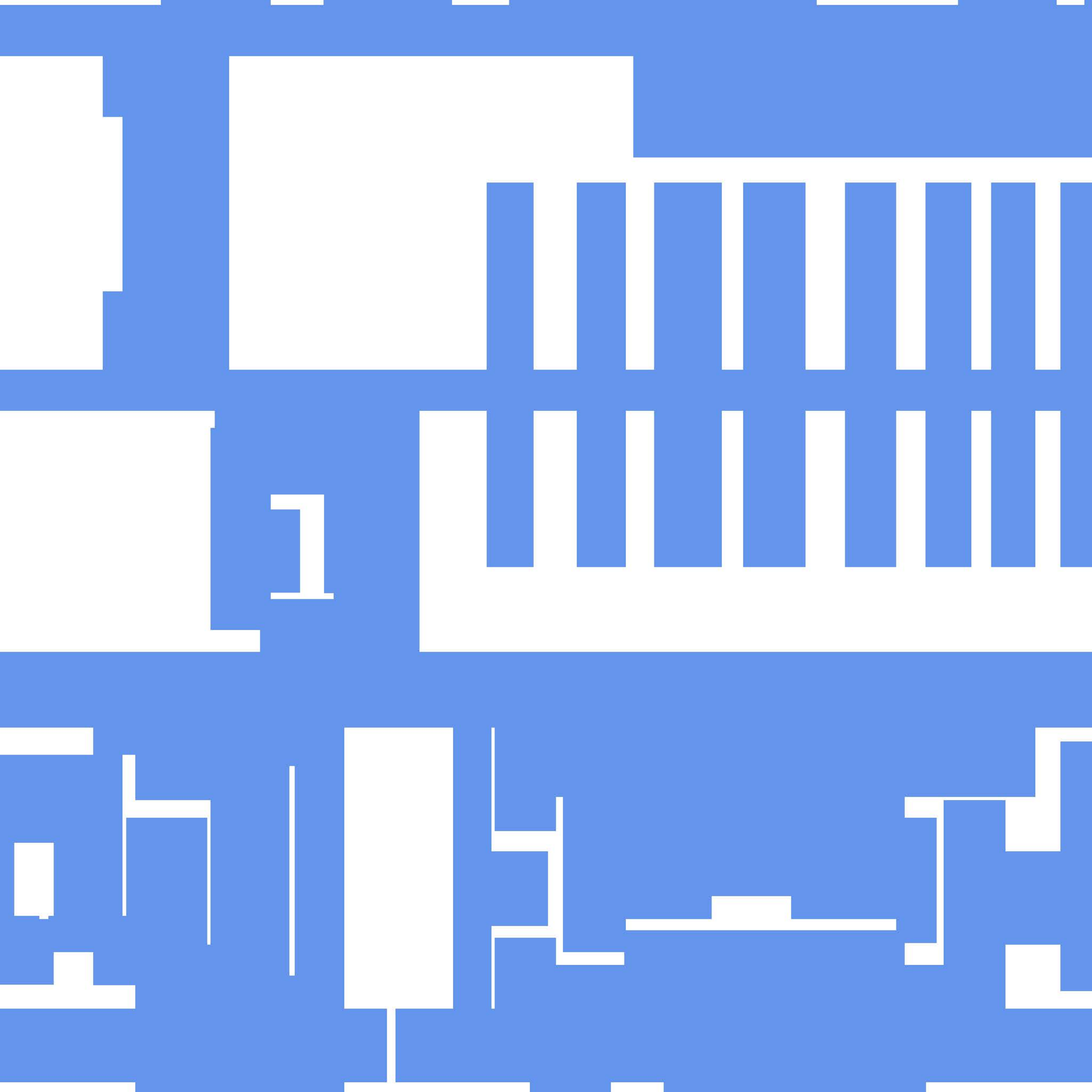} }
    \subfloat[]{ \includegraphics[width=0.30\linewidth]{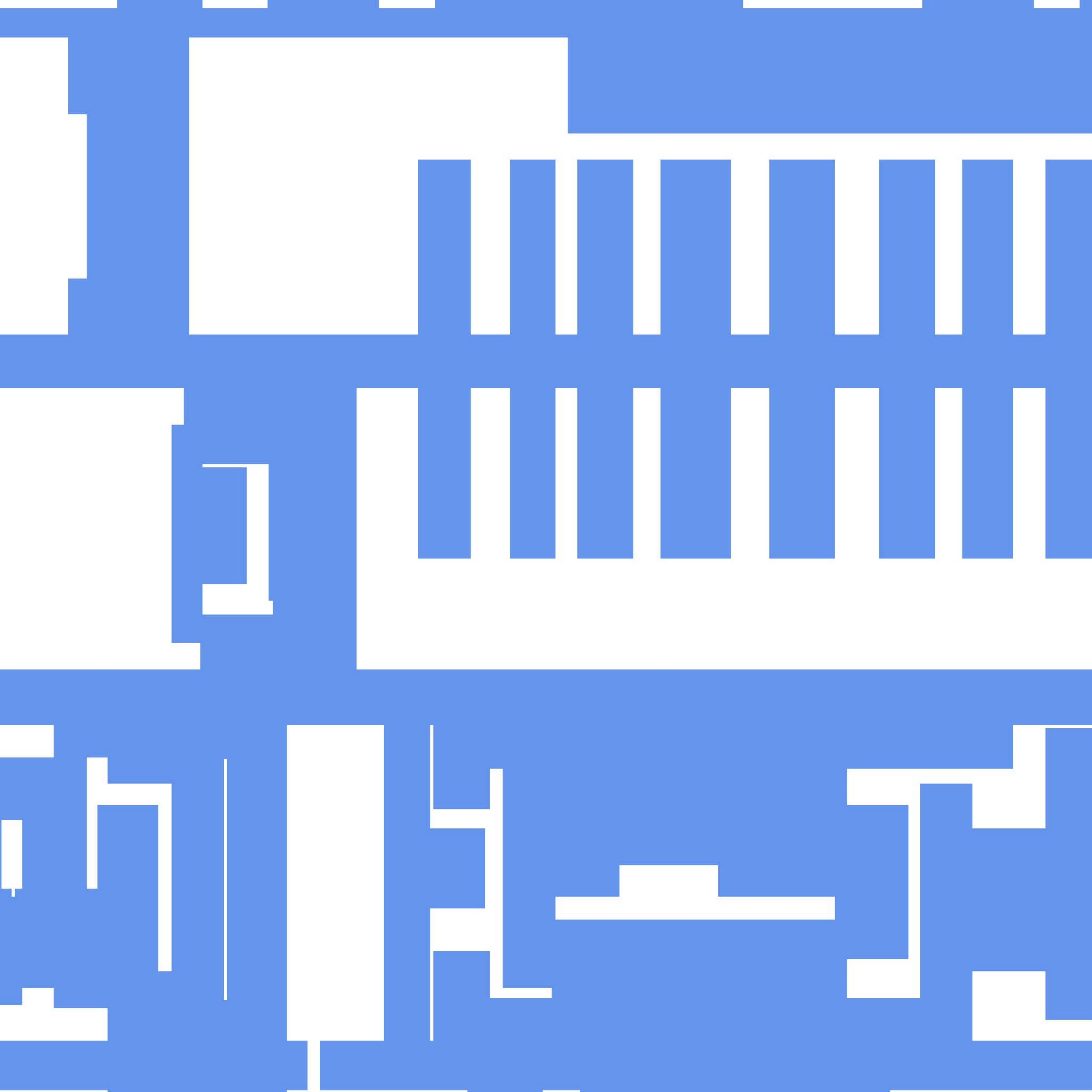} }
    \subfloat[]{ \includegraphics[width=0.30\linewidth]{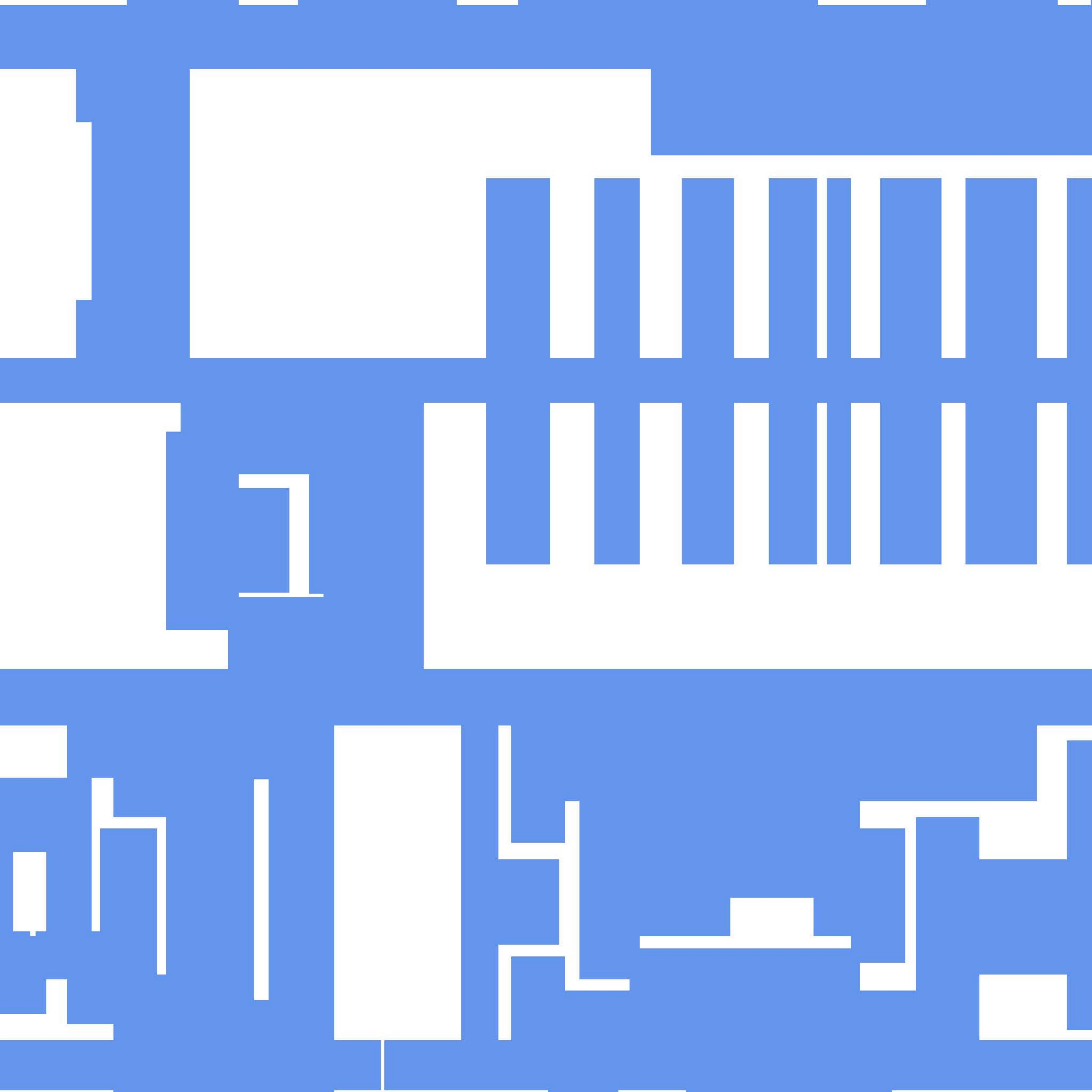} }
    \caption{Different layout patterns that are generated from a single topology with the same design rule. }
    \label{fig:single}
\end{figure}

\begin{figure}[tb!]
    \centering
    \subfloat[Normal rule]{ \includegraphics[width=0.30\linewidth]{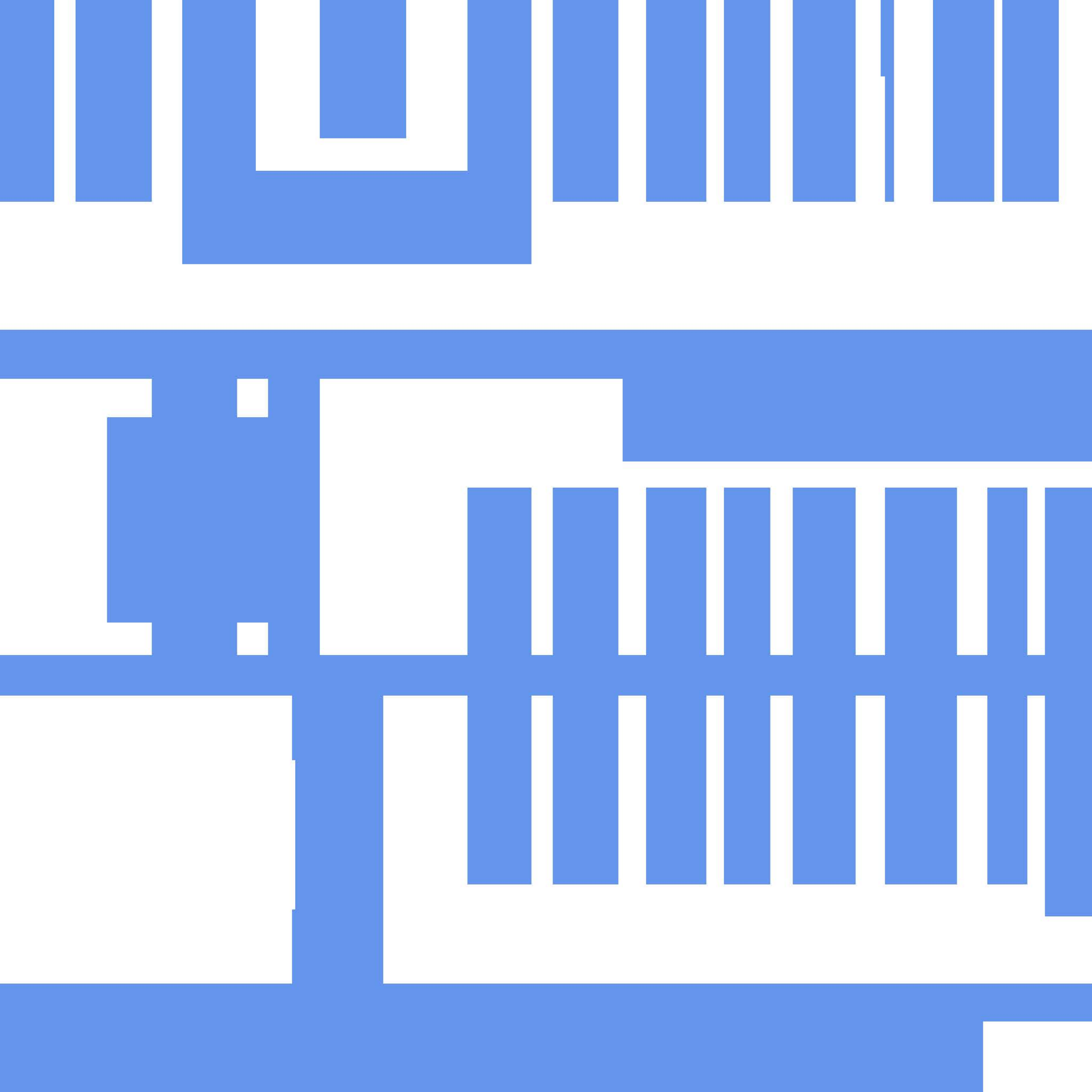} }
    \subfloat[Larger $\textit{Space}_\textit{min}$]{ \includegraphics[width=0.30\linewidth]{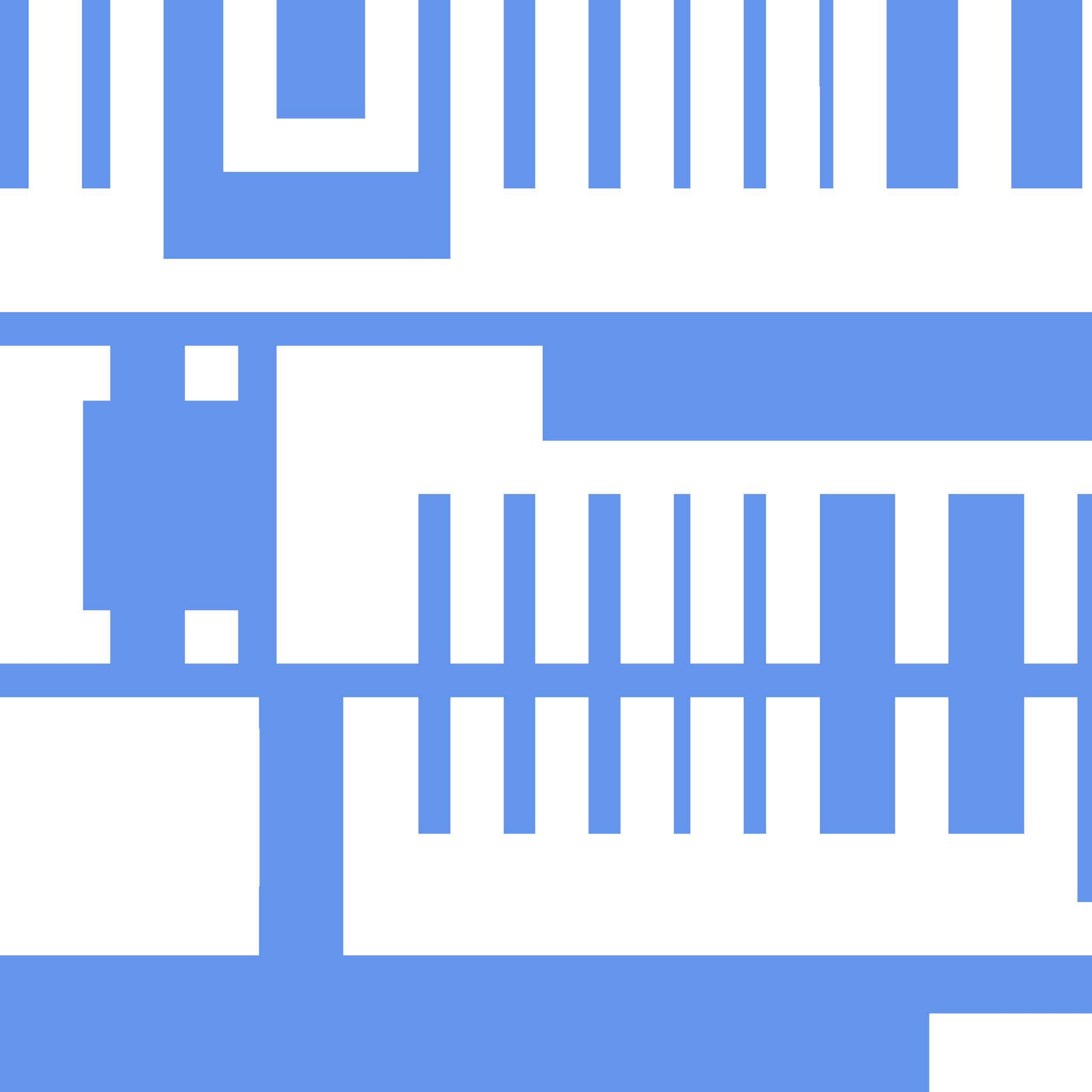} }
    \subfloat[Smaller $\textit{Area}_\textit{max}$]{ \includegraphics[width=0.30\linewidth]{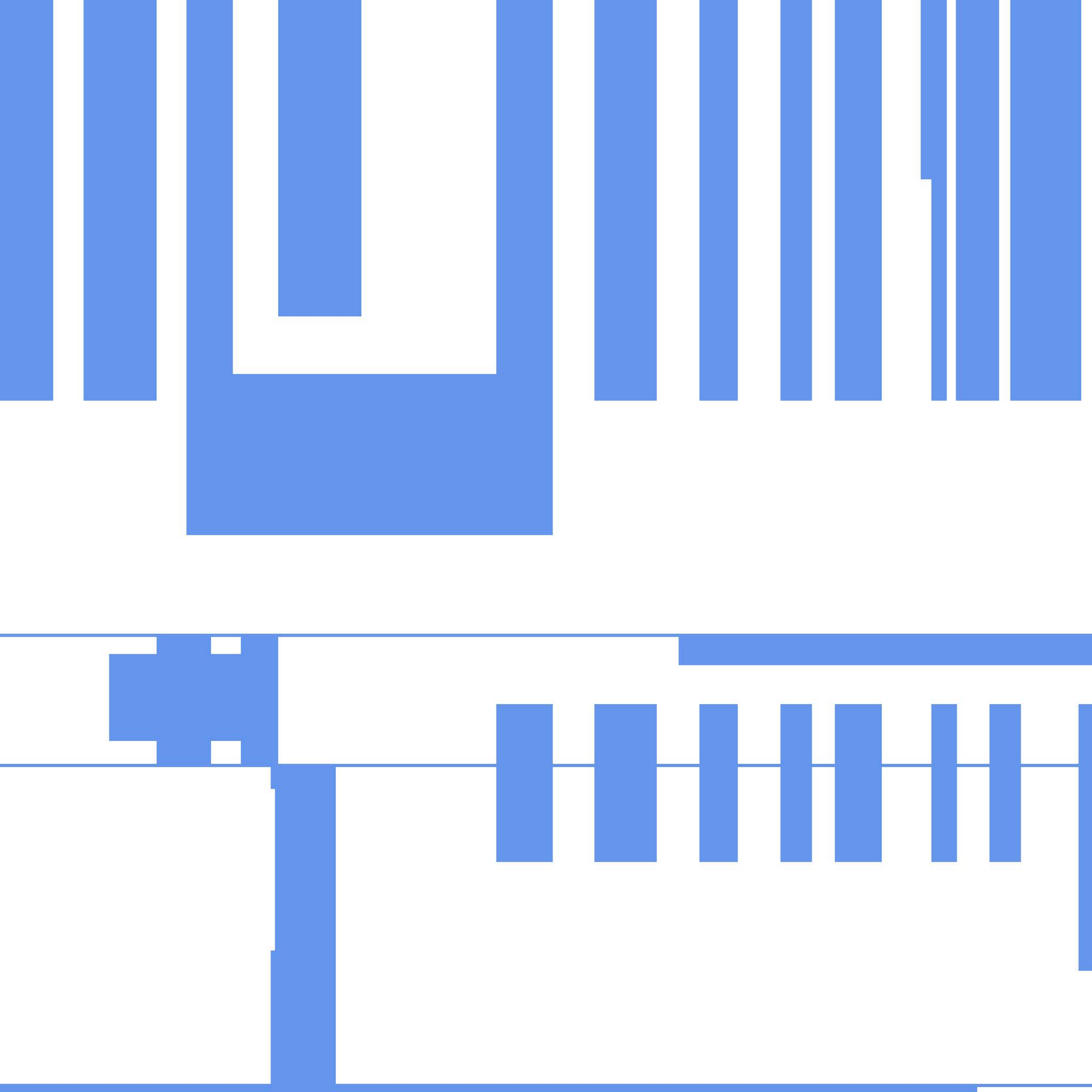} } \\
    \subfloat[Normal rule]{ \includegraphics[width=0.30\linewidth]{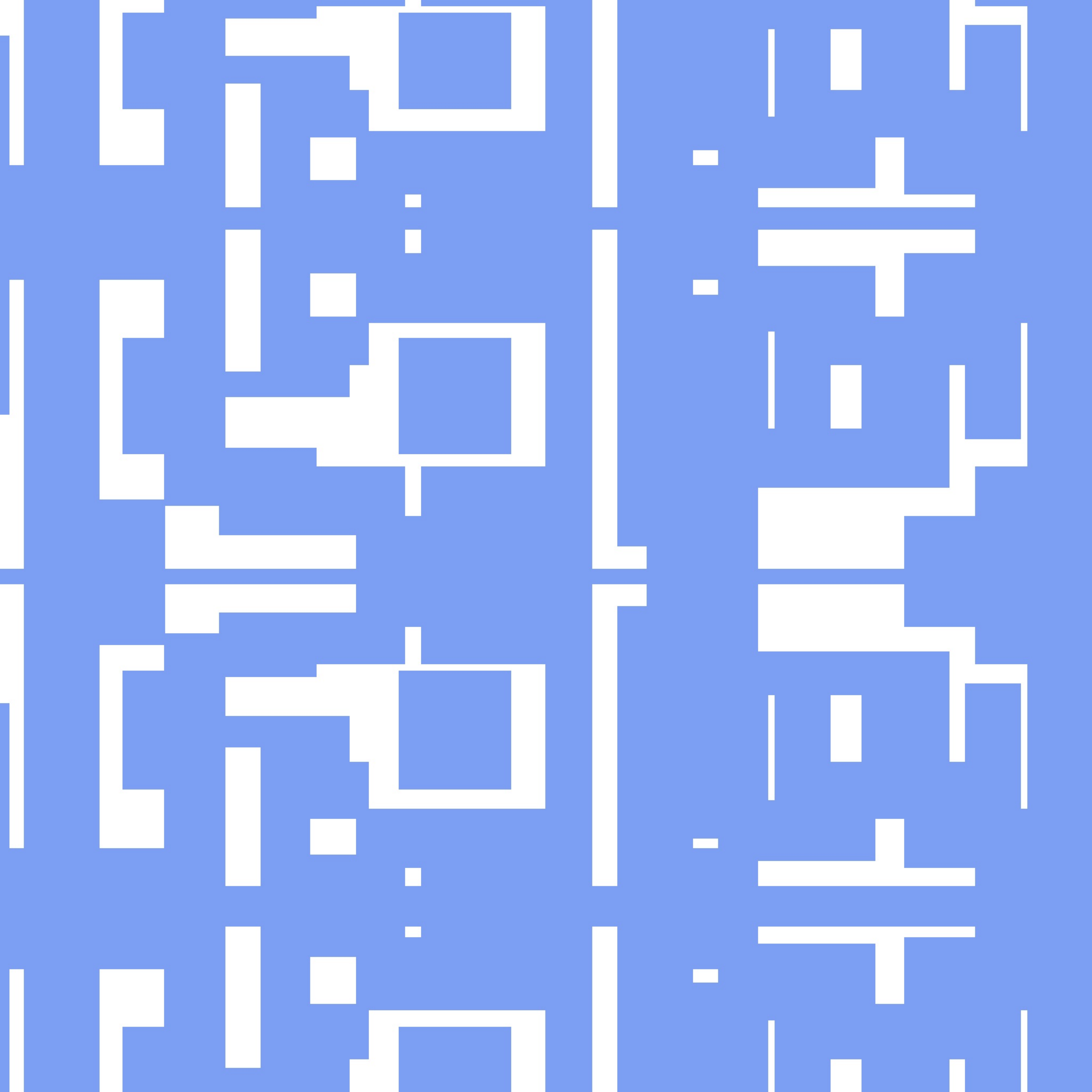} }
    \subfloat[Larger $\textit{Width}_\textit{min}$]{ \includegraphics[width=0.30\linewidth]{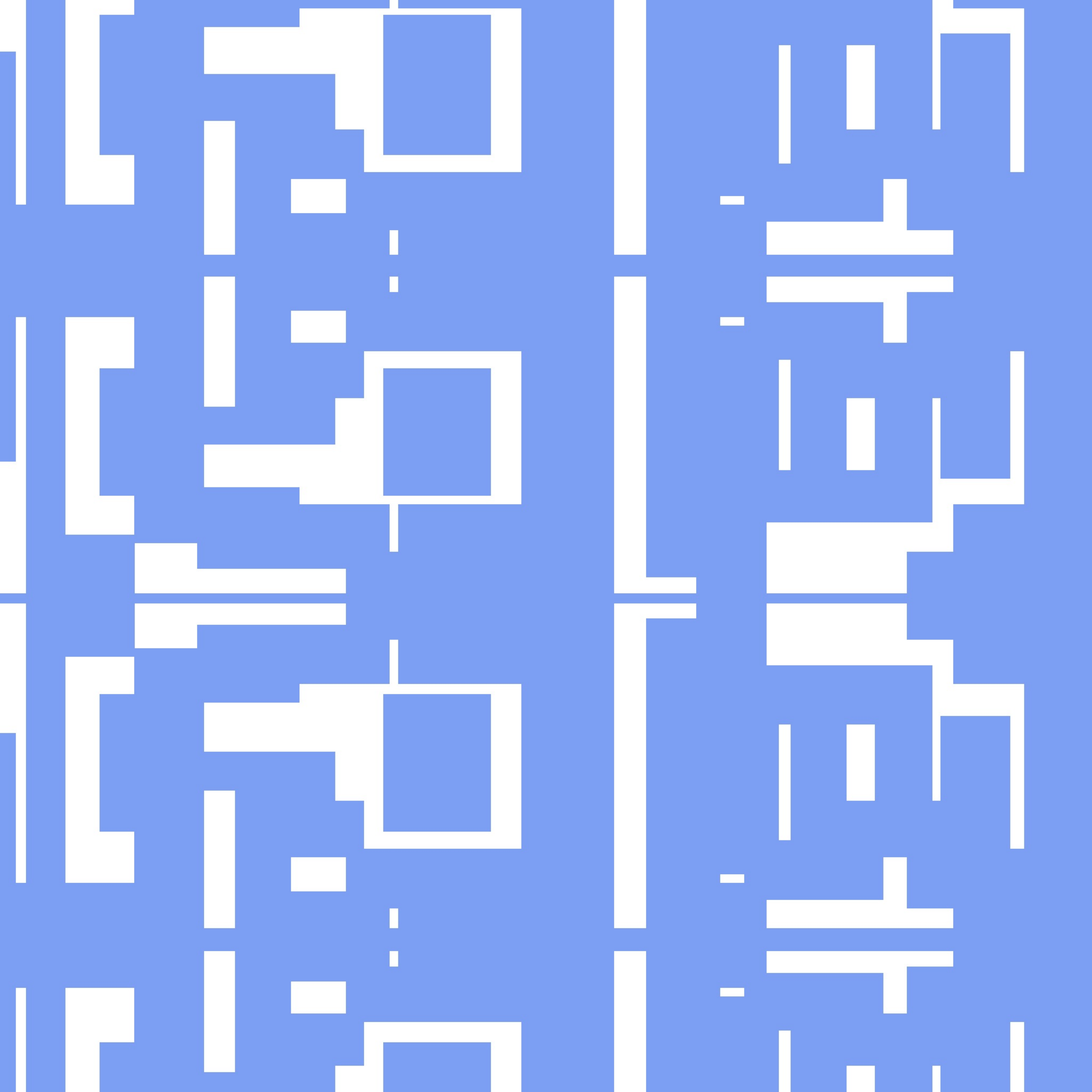} }
    \subfloat[Smaller $\textit{Area}_\textit{max}$]{ \includegraphics[width=0.30\linewidth]{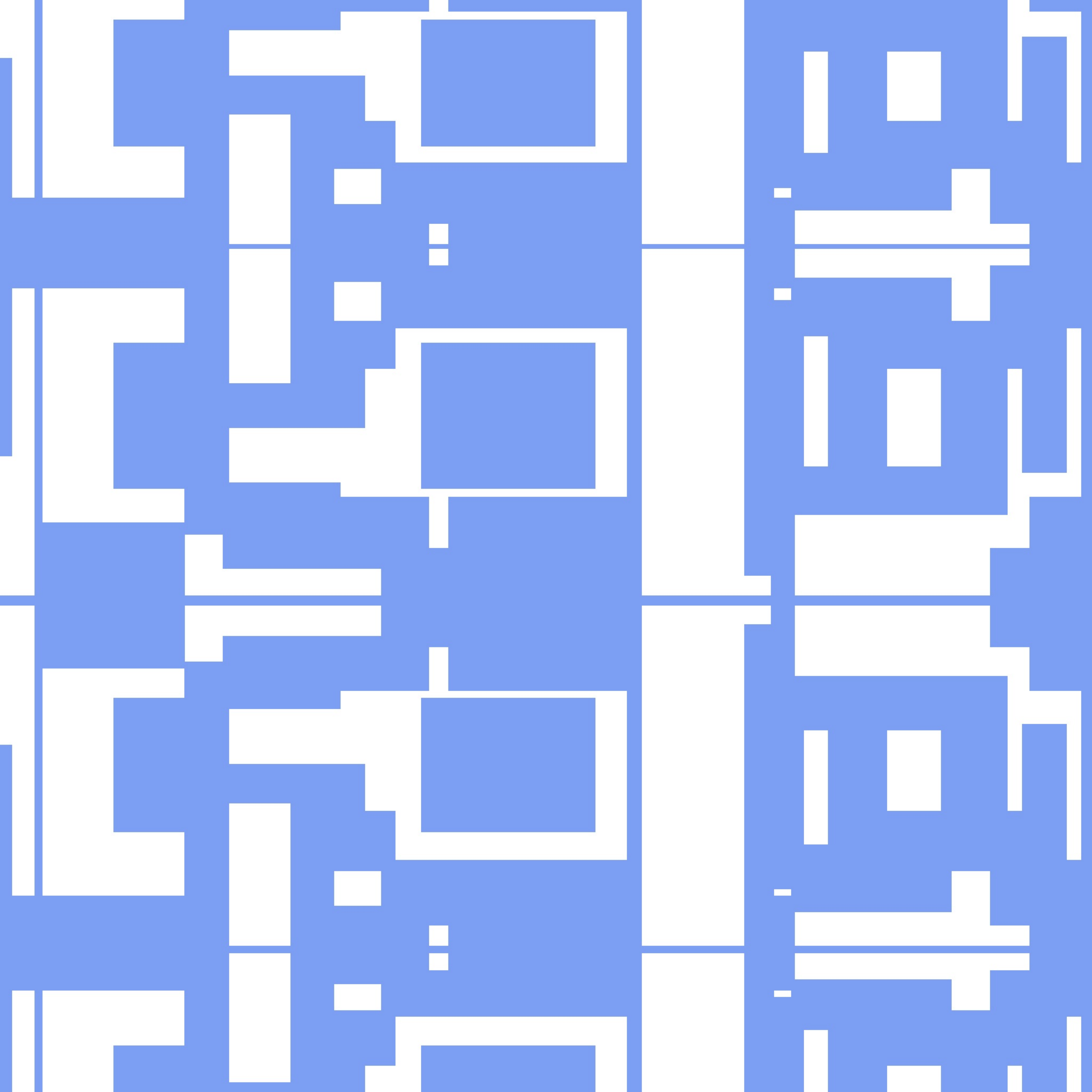} }
    \caption{
        Layout patterns that are generated from the same topology with different design rules.
    }
    \label{fig:rules}
\end{figure}

\minisection{Diffusion Model Configuration.} In line with previous research \cite{ho2020denoising, austin2021structured}, we adopt a U-Net architecture \cite{ronneberger2015u} as the foundation of our discrete diffusion model to approximate the posterior distribution during the reverse diffusion process. The model operates at four different feature map resolutions: $[32\times32, 16\times16, 8\times8, 4\times4]$. Each resolution stage consists of two convolutional residual blocks, with the number of convolutional channels set to $\left[128, 256, 256, 256\right]$, respectively. At the 16$\times$16 resolution stage, a self-attention block is inserted between the two convolutional blocks. Moreover, the time step $k$ is embedded into each residual block using sinusoidal positional encodings \cite{vaswani2017attention}. To guarantee convergence of the forward diffusion process to a uniform stationary distribution, we set the number of diffusion steps $K$ to 1000. The noise schedule $\beta_k$ increases linearly from $0.01$ to $0.5$. During inference, we found that reversing the process in $m=10$ steps offers an effective trade-off between pattern quality and computational efficiency, as discussed in \Cref{sec:2.6}.

\minisection{Training Details.} The diffusion model is trained for $0.5$M iterations with a batch size of 128, using a learning rate of $2e$-4 and the Adam optimizer. The following hyperparameters are used: a dropout rate of $0.1$, gradient clipping set to $1$, and a loss coefficient $\lambda$ of $0.001$. As described in \Cref{sec:2.5}, data augmentation is applied randomly during training using reliable augmentation techniques. The probabilities for the augmentation methods—\textit{Random Flip}, \textit{Random Rotation}, \textit{Symmetric Mirror}, and \textit{Concatenate and Crop}—are $[0.5, 1.0, 0.5, 0.5]$. The training process spans approximately 20 hours on 8 NVIDIA RTX 3090 GPUs.

\begin{figure*}[tb!]
    \centering
    \includegraphics[width=.9\linewidth]{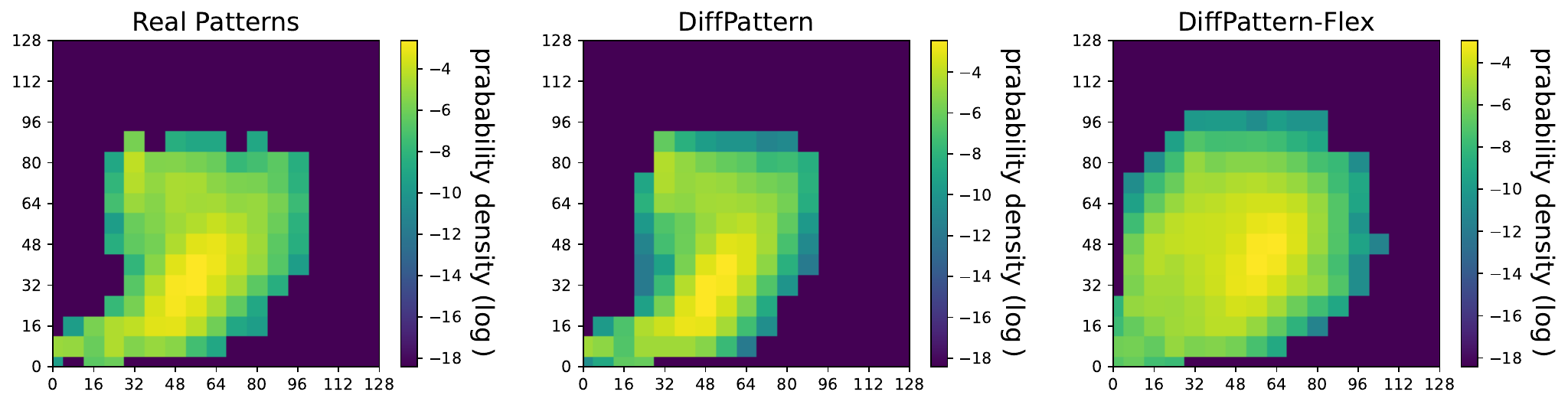}
    \caption{An illustration of complexity distribution. Reliable pattern augmentation enriches the pattern distribution while maintaining the reasonability of the enhanced patterns.}
    \label{fig:complexity}
\end{figure*}

\subsection{Pattern Diversity and Legality}
\label{sec:4.2}

To evaluate the models \cite{theis2015note, wang2023evaluation}, we calculate the diversity of the generated patterns using \Cref{eq:diveristy}, while the legality of the patterns is verified with the tool \textit{Klayout}, which checks compliance with the design rules as outlined in \Cref{sec:2.3}.

For a fair comparison with previous methods, we randomly synthesize 100,000 topologies, assigning each generated topology a pair of geometric vectors during the legality assessment phase. In this evaluation, we refer to our method as \tool{DiffPattern}-Flex. It is worth noting that \tool{DiffPattern}-Flex can generate a large number of legal patterns for each topology, as the solution to the optimization problem in \Cref{eq:nonlinear} is not unique.

We compare \tool{DiffPattern}-Flex with several learning-based layout pattern generation approaches. Specifically, CAE \cite{yang2019deepattern} represents a standard convolutional autoencoder model, while VCAE \cite{zhang2020layout} utilizes a variational convolutional autoencoder. Both of these methods are pixel-based. LegalGAN \cite{zhang2020layout} is a post-processing approach that legalizes generated topologies by making necessary modifications. In contrast, LayouTransformer \cite{wen2022layoutransformer} employs a sequence-based transformer model to synthesize new sequential representations of layout patterns, bypassing direct topology generation. Lastly, DiffPattern \cite{wang2023diffpattern}, our primary competitor, uses a discrete diffusion model to generate topology matrices.
{We also implemented the rule-based pattern generation method in \cite{ye2019lithoroc}. An original pattern is split into four sub-patterns. The sub-patterns are further randomly flipped and rotated and form an enhanced sub-pattern library. A new pattern consists of four randomly picked sub-patterns from the library. We also generate 100,000 patterns in this way and test the diversity and legality of the generated patterns.}

As presented in \Cref{tab:diversity}, \tool{DiffPattern}-Flex outperforms other methods in terms of legality, achieving a perfect performance ({\it i.e.},~100\%) under standard conditions. This is due to the topology pre-filtering and rule-based 2D legal pattern assessment. Furthermore, \tool{DiffPattern}-Flex demonstrates a marked improvement in diversity (from 10.815 to 11.713) over DiffPattern, thanks to its pattern augmentation techniques. The robustness of the generation and assessment methods ensures the reasonableness of the augmented patterns.

Illustrative examples generated by \tool{DiffPattern}-Flex are shown in \Cref{fig:generated}, highlighting the diversity of the patterns produced. These examples underscore the capability of our method to generate detailed, reasonable patterns that maintain a high degree of diversity across the dataset.

Moreover, the legality of patterns produced by \tool{DiffPattern}-Flex is consistently ensured through its rule-based legal pattern assessment and topology pre-filtering process. When design rules are modified, our method allows users to quickly generate a new batch of diverse, compliant patterns without retraining the topology generation model. This flexibility is explored further in the next subsection.

\subsection{Flexibility}
\label{sec:4.3}

A key advantage of \tool{DiffPattern}-Flex is the flexibility provided by its white-box 2D legal pattern assessment phase. Below, we illustrate two applications that leverage this flexibility.

\minisection{Generating Multiple Patterns from a Single Topology.} Given a set of design rules and a specific topology, the non-linear system in \Cref{eq:nonlinear} often admits numerous legal solutions, {\it i.e.,} different geometric vector pairs. Each legal solution corresponds to a valid layout pattern, and these patterns, while differing in geometric vectors, share the same underlying topology. Such variations can be valuable for certain downstream tasks. \Cref{fig:single} demonstrates examples where multiple layout patterns are generated from a single topology by using different geometric vectors.

\minisection{Generating Legal Patterns under Varying Design Rules} 
{The decoupling of legalization and topology generation in \tool{DiffPattern}-Flex enables the flexibility of design rule modification. Since changes in design rules do not affect the distribution of topology matrices, the trained topology generation model remains applicable across varying design rules. \Cref{fig:rules} presents examples where multiple layout patterns are generated from a single topology, each adhering to a distinct set of design rules.

In this paper, most cases can be resolved within a reasonable time budget using the design rules defined in \Cref{sec:2.3}. The experimental results demonstrate that the proposed legalization method can handle changes in the constants of the design rules. However, it is important to note that in scenarios where the design rules are extremely strict and complex, the generated topology matrices may fail to find a legal solution within a limited time using the proposed legalization methods. In such cases, advanced legalization techniques should be developed to reduce computational costs, which we leave for future work.}

 \begin{table*}[tb!]
    \centering
    \caption{Model efficiency of \tool{DiffPattern}-Flex under different accelerator factor $m$. Nvidia RTX 3090 GPU is used for topology sampling in this table. }
    \label{tab:efficiency-sample}
        \begin{tabular}{c|cccc}
            \toprule
            Method & $m$& Cost Time (s)& Acceleration &Diversity \\
            \midrule
            Sampling &1&0.544 &1.00$\times$ & 11.724\\    
            {Fast-Sampling} & {5} & {0.118} & {4.61$\times$}& {11.715} \\
            Fast-Sampling & 10 & 0.065 & 8.37$\times$ & 11.713 \\
            {Fast-Sampling} & {20} & {0.039} & {13.95$\times$} & {10.573} \\
            \bottomrule 
        \end{tabular}
\end{table*}

\begin{table*}[tb!]
    \centering
    \caption{Model efficiency of \tool{DiffPattern}-Flex. Intel(R) Xeon(R) Gold 6326 CPU @ 2.90GHz is used to figure out the non-linear system in this table. }
    \label{tab:efficiency-solve}
    
        \begin{tabular}{c|ccc}
            \toprule
            Method & Extra Information &Cost Time (s)& Acceleration  \\
            \midrule
            Solving-R & Not required & 0.269&1.00$\times$\\
            Solving-E & Required&0.117&2.30$\times$\\
            Solving-D & Not required & 0.108&2.48$\times$ \\
            
            \bottomrule 
        \end{tabular}
\end{table*}

\subsection{Distribution of Complexity}

Diversity is a vital metric for assessing the quality of the generated pattern library. As outlined in \Cref{sec:2.3}, diversity is quantified using the Shannon entropy of the distribution of pattern complexity, \textit{i.e.}, the number of scan lines that intersect a pattern along both the x-axis and y-axis. The distribution of complexity is visualized in \Cref{fig:complexity}.

The patterns generated by DiffPattern exhibit a complexity distribution comparable to that of real-world patterns. However, due to the robust pattern augmentation techniques, the patterns generated by \tool{DiffPattern}-Flex show increased diversity.
The heatmap generated by \tool{DiffPattern}-Flex covers regions that are missing from the heatmap of real patterns, suggesting that \tool{DiffPattern}-Flex is capable of discovering novel combinations of existing patterns.

This visualization underscores our capability to produce high-quality, diverse layout patterns.





\subsection{Model Efficiency}

Efficiency is a critical measure when evaluating methods for layout generation. In our approach, since the processes of topology generation and layout pattern validation are decoupled, we separately record the average time taken to sample a new topology and to solve for a legal solution of \Cref{eq:nonlinear}. The corresponding results are provided in \Cref{tab:efficiency-sample,tab:efficiency-solve}.

As outlined in \Cref{sec:2.6}, the sampling procedure can be accelerated by approximately a factor of $m$, where $m$ represents the number of backward steps executed during each model inference. In our implementation, we set $m=10$, and the findings show that the sampling process can be sped up by $8.37\times$ with only a slight reduction in the diversity of the generated patterns.

\begin{figure}[tb!]
    \centering
    \subfloat[Topology 1]{ \includegraphics[width=0.30\linewidth]{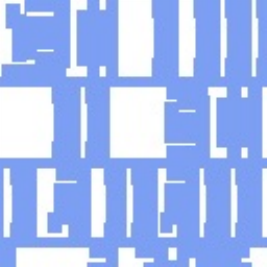} }
    \subfloat[Topology 2]{ \includegraphics[width=0.30\linewidth]{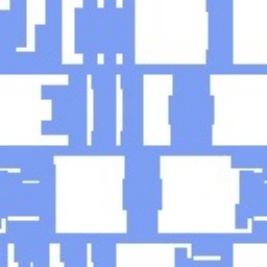} }
    \subfloat[Topology 3]{ \includegraphics[width=0.30\linewidth]{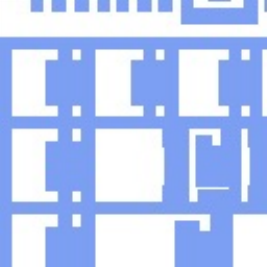} } \\
    \subfloat[Initial w/ Solving-E]{ \includegraphics[width=0.30\linewidth]{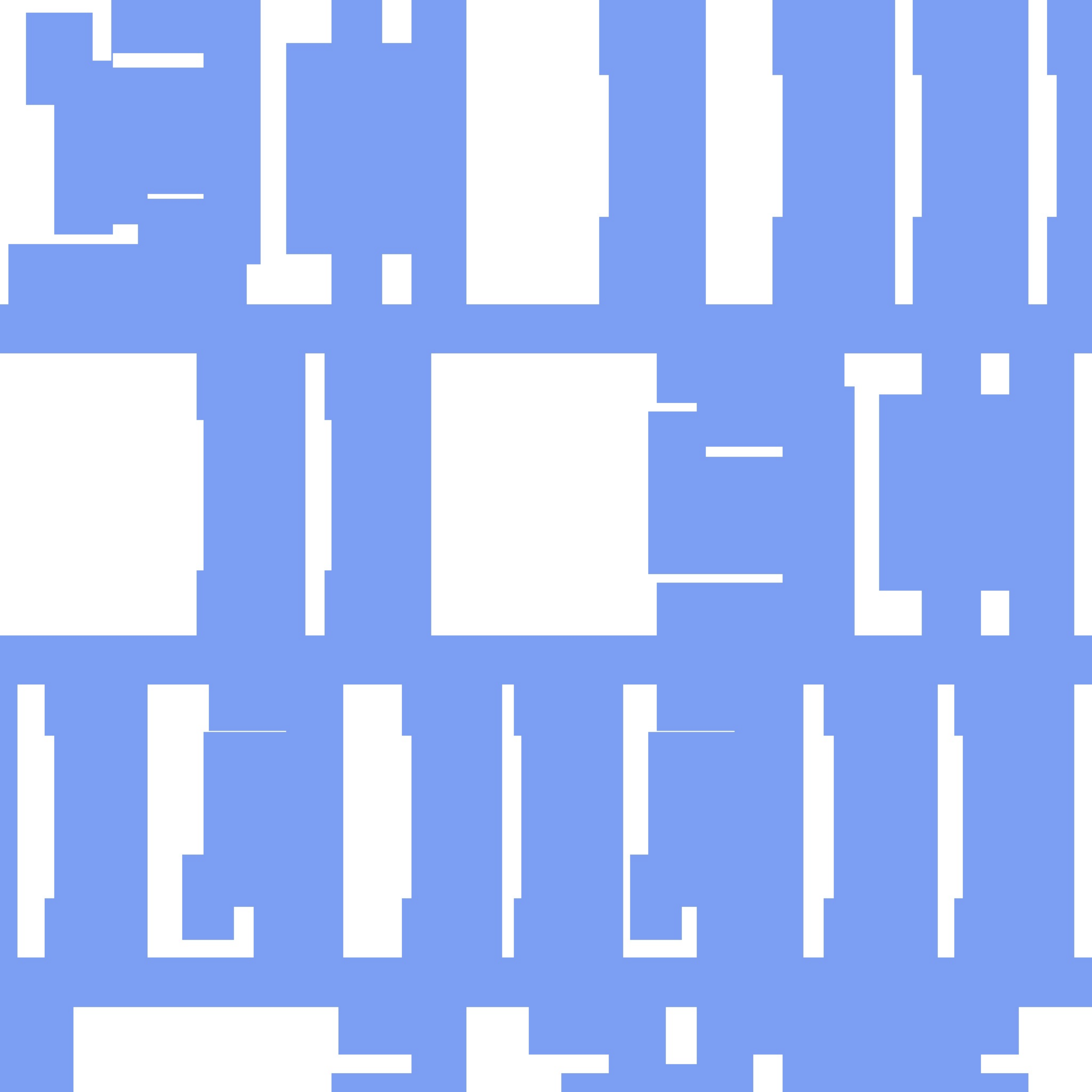} }
    \subfloat[Initial w/ Solving-E]{ \includegraphics[width=0.30\linewidth]{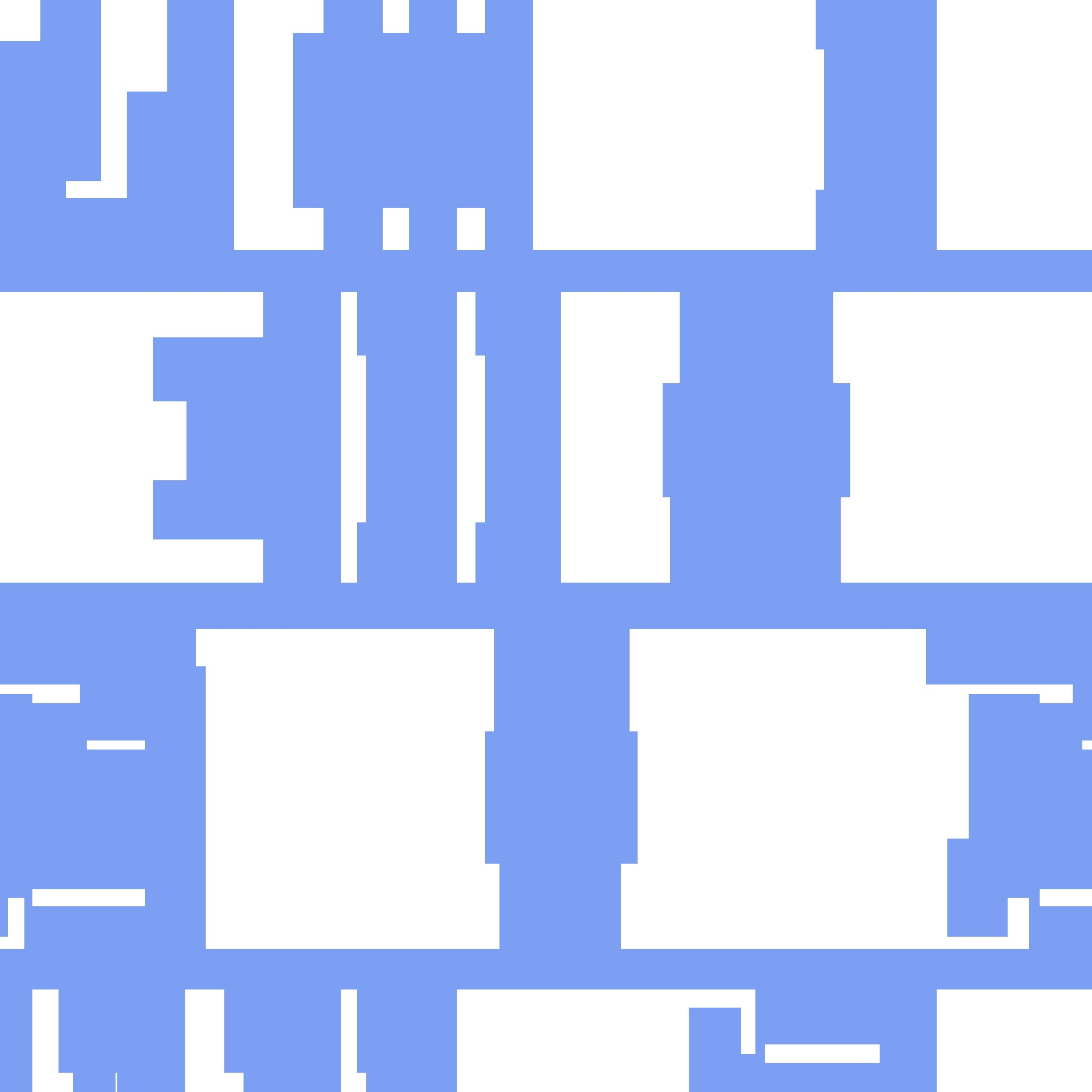} }
    \subfloat[Initial w/ Solving-E]{ \includegraphics[width=0.30\linewidth]{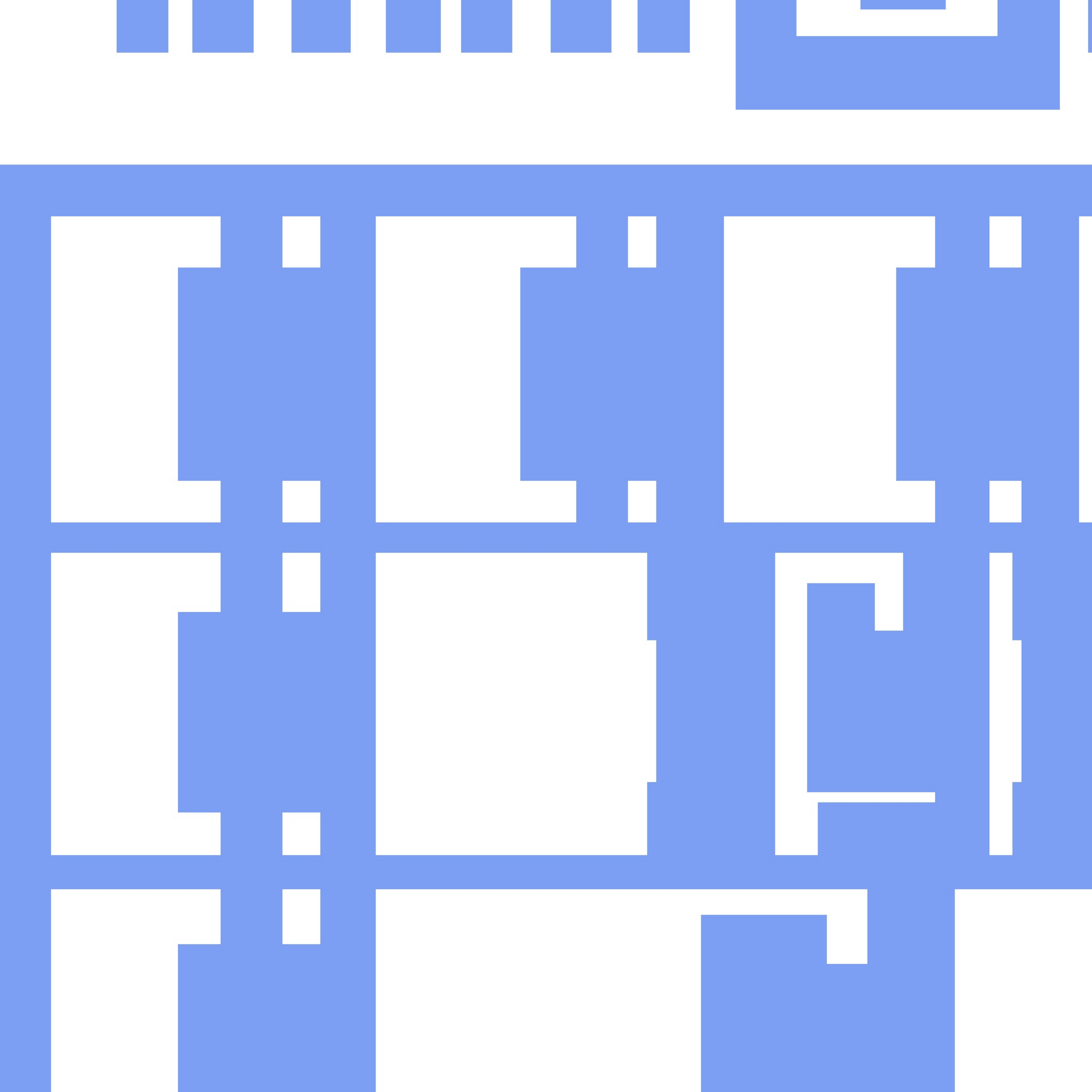} } \\
    \subfloat[Initial w/ Solving-D]{ \includegraphics[width=0.30\linewidth]{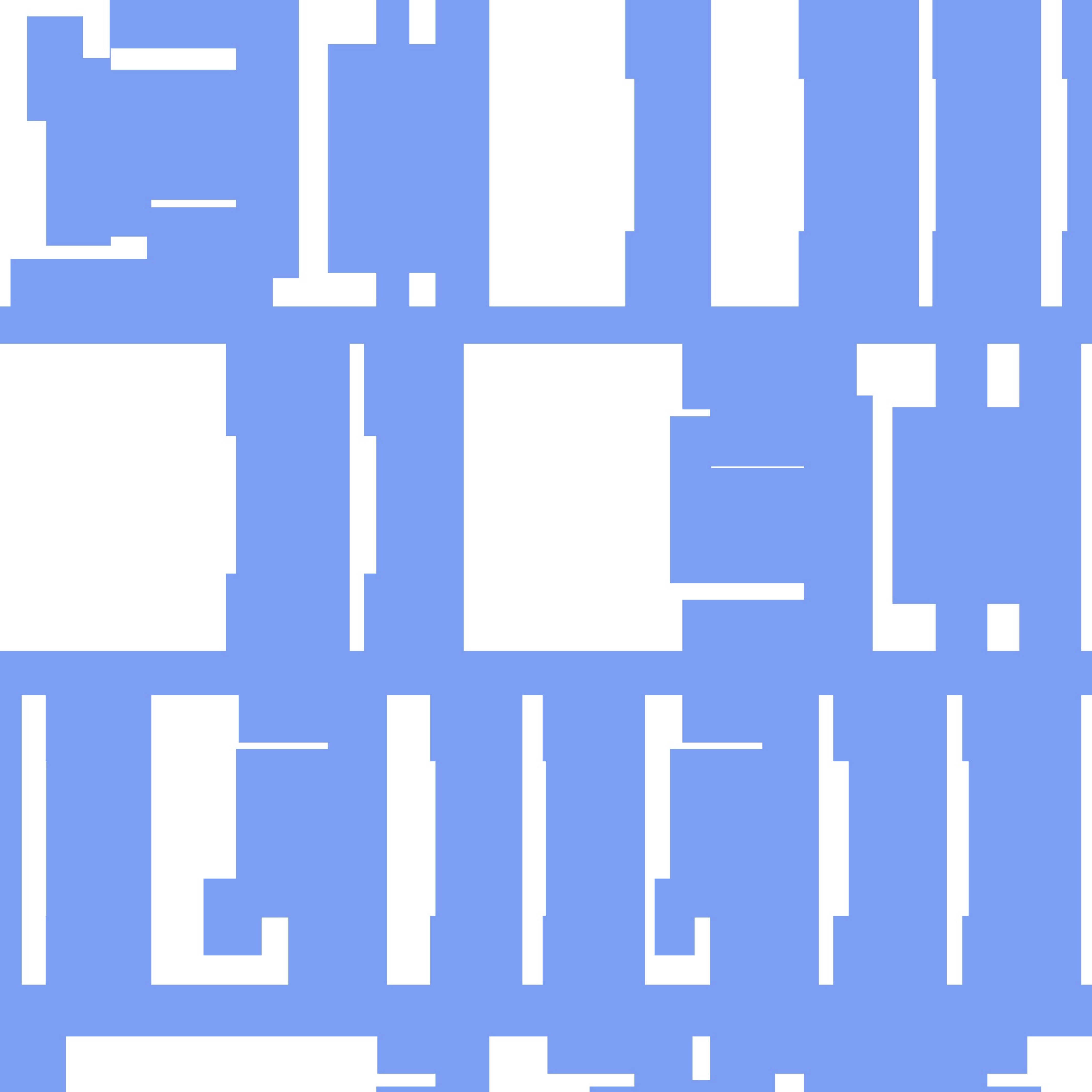} }
    \subfloat[Initial w/ Solving-D]{ \includegraphics[width=0.30\linewidth]{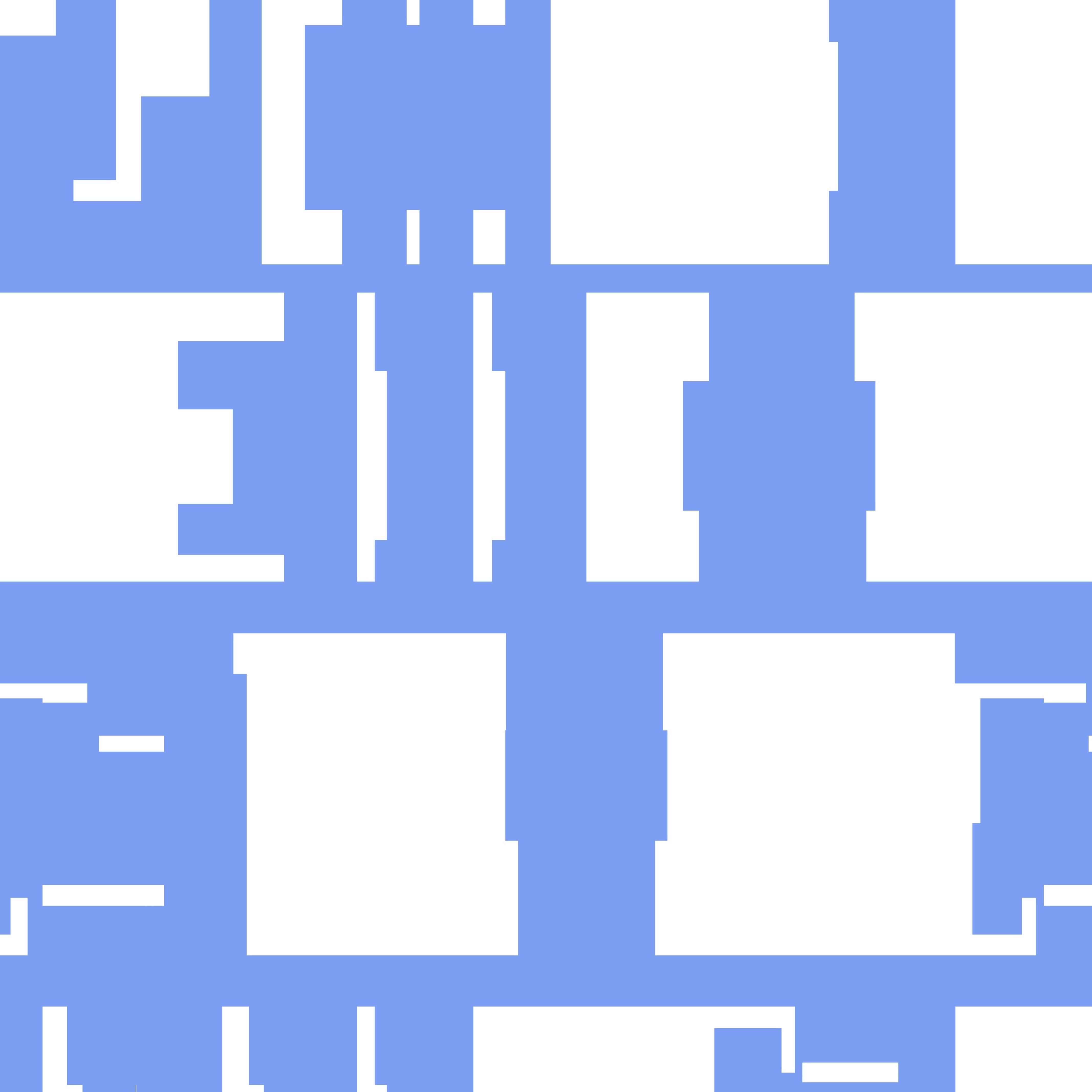} }
    \subfloat[Initial w/ Solving-D]{ \includegraphics[width=0.30\linewidth]{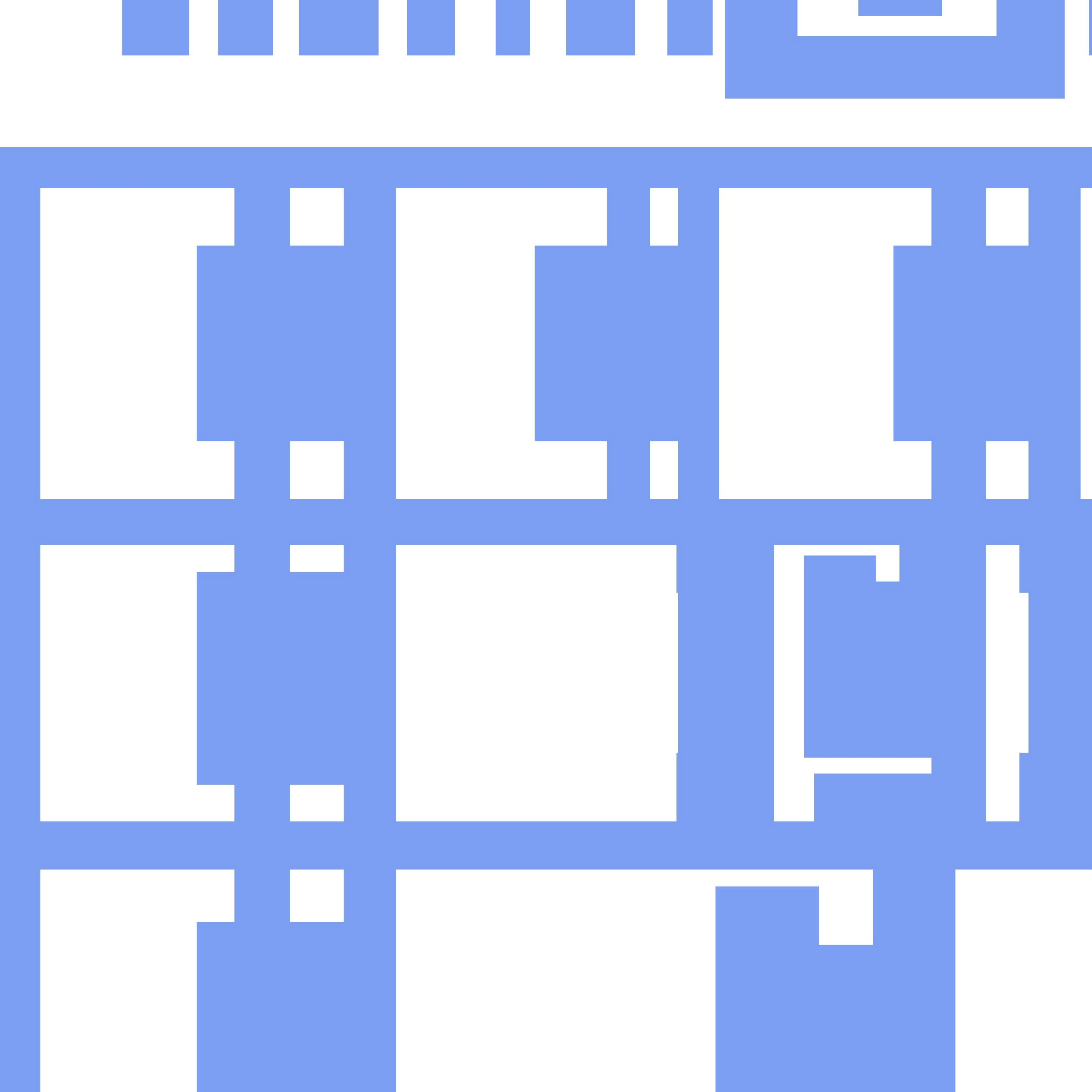} } \\
    \subfloat[Final Result 1]{ \includegraphics[width=0.30\linewidth]{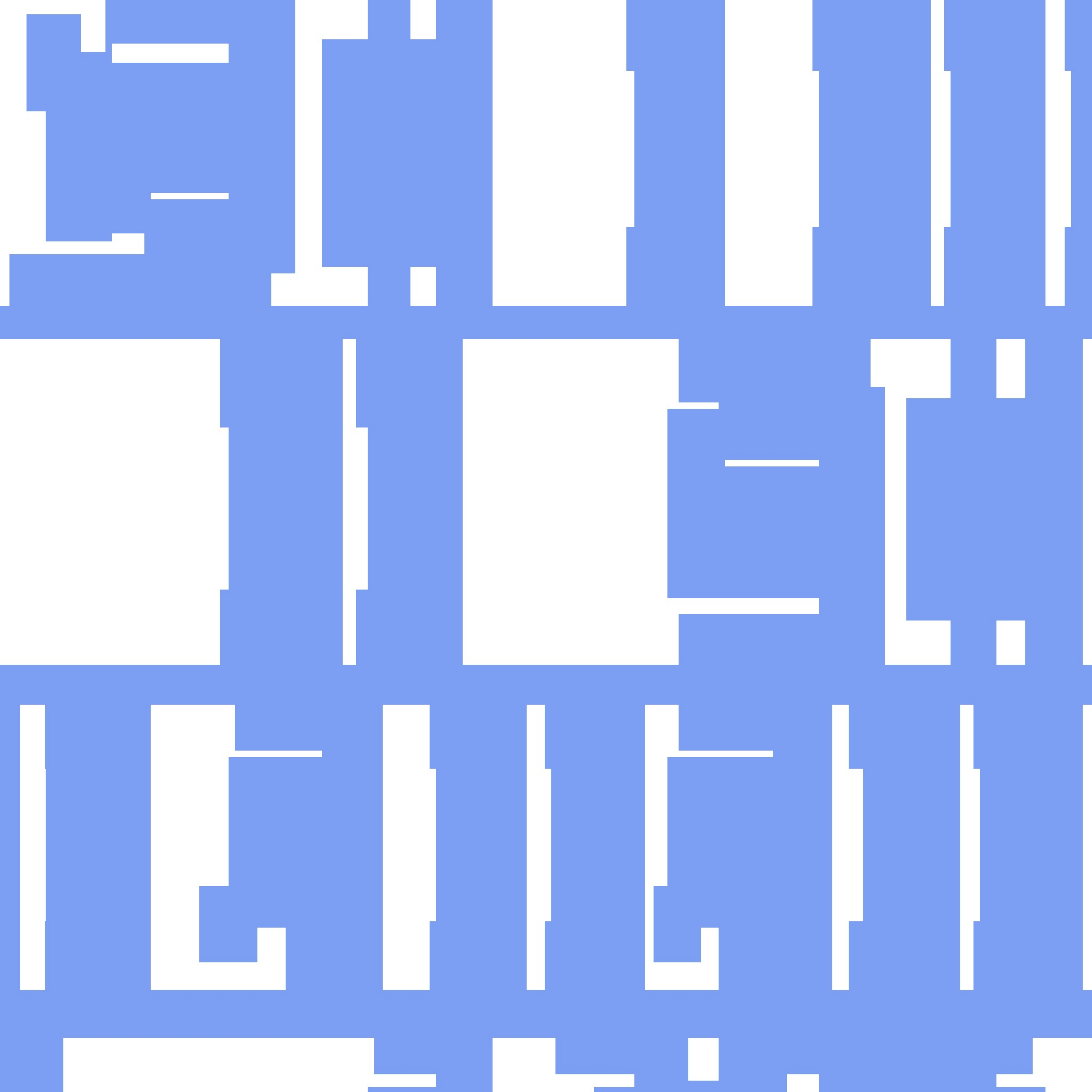} } 
    \subfloat[Final Result 2]{ \includegraphics[width=0.30\linewidth]{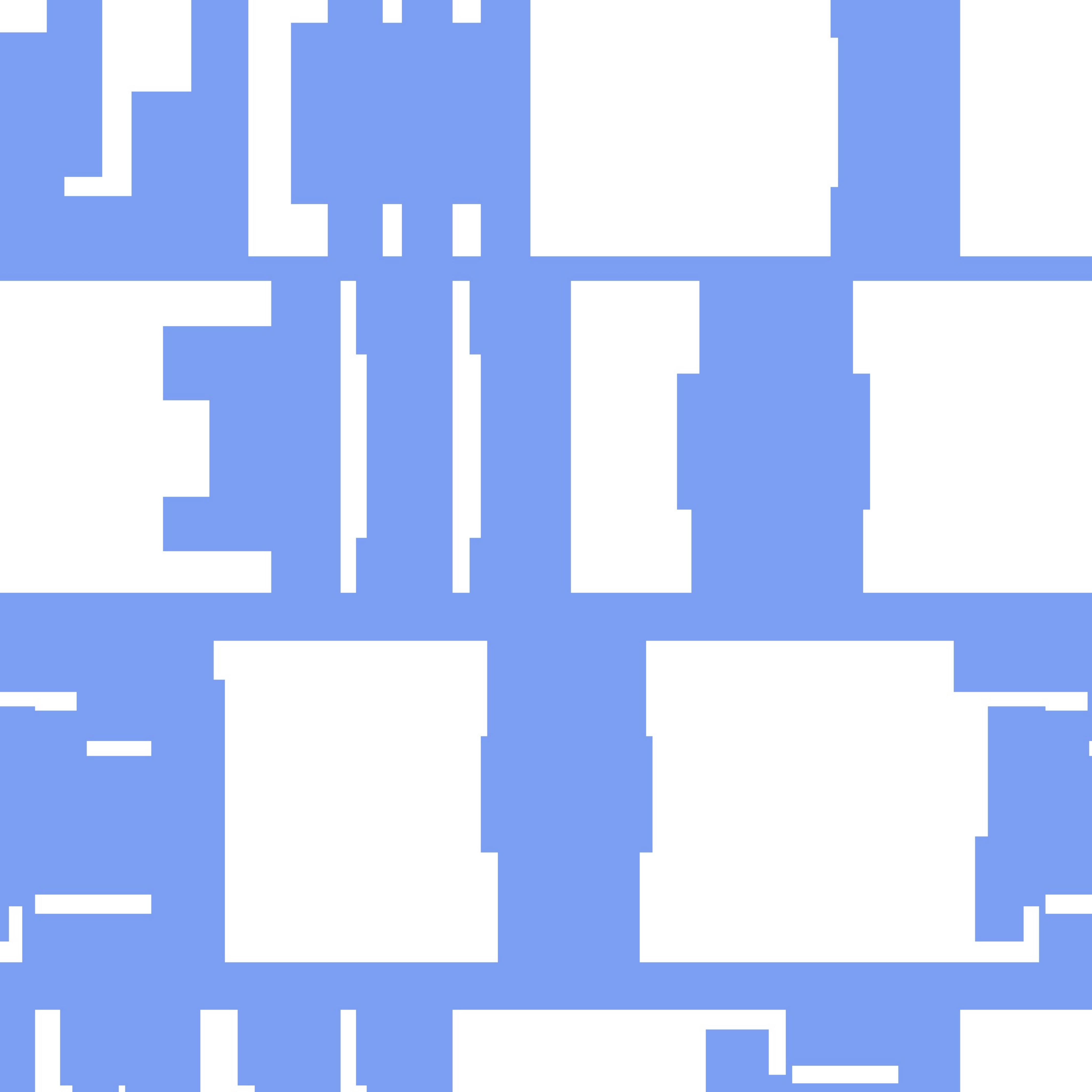} } 
    \subfloat[Final Result 3]{ \includegraphics[width=0.30\linewidth]{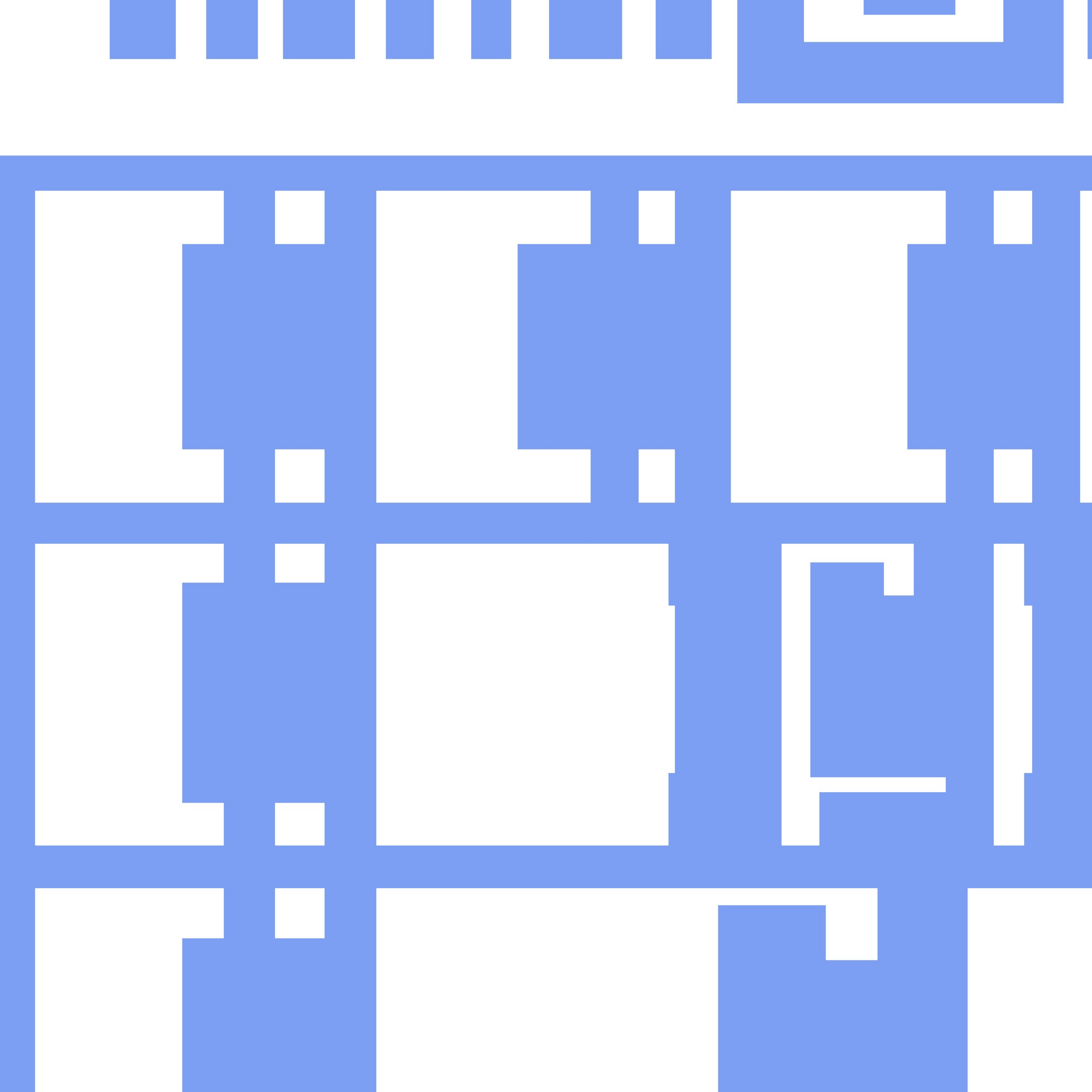} } \\
    \caption{
        Illustration of initialization and final solution in the legalization phase. 
    }
    \label{fig:initial}
\end{figure}

In scenarios where both the target and source patterns adhere to the same design rules, as mentioned in \Cref{sec:2.4}, an existing pair of geometric vectors can be randomly chosen to initialize the nonlinear system, which significantly speeds up the process. This method is referred to as \textit{Solving-E}. When compared to the version using random initialization, termed \textit{Solving-R}, \textit{Solving-E} offers an average acceleration of $2.30\times$. Several examples illustrating the outcomes of \textit{Solving-E} are shown in \Cref{fig:initial}.

In scenarios where the target pattern and source pattern follow different design rules or when suitable geometric vectors are unavailable, we can accelerate the solving process by providing a better initialization using the divide method described in \Cref{sec:2.6}. This version is denoted as \textit{Solving-D}. Our experimental results show that \textit{Solving-D} achieves an average acceleration of $2.48\times$ in our cases. The total time for \textit{Solving-D} includes both the solution of sub-problems and the main problem. The divide size in our implementation is fixed as $4$, which means we divide the hard main problem into four sub-problems for each tensor matrix. We also show some cases in \Cref{fig:initial}. In the major cases, the final solution has a mirror difference compared with the initialization obtained from sub-problems.

\begin{figure}
    \centering
    \includegraphics[width=0.8\linewidth]{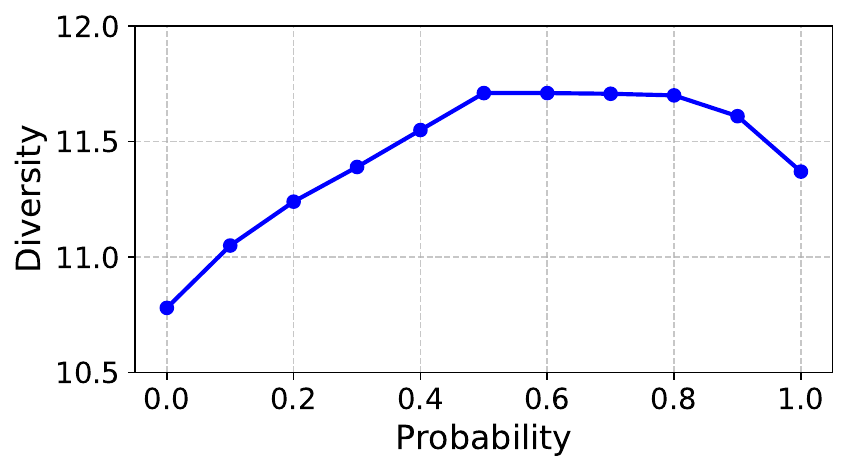}
    \caption{{The effect of different probabilities of concatenate and crop augmentation.}}
    \label{fig:abla-pro}
\end{figure}

{
\subsection{Ablation Study}
\label{sec:abla}

In this subsection, we conduct some ablation studies on how each part in \tool{DiffPattern}-Flex affects the results.

\minisection{Comparison with Continual Diffusion Model.} 
In the main part of this paper, we utilize a discrete diffusion model to directly generate a discrete topology matrix, fully leveraging the capacity of the neural network. Here, we compare our method with a continuous diffusion baseline~\cite{ho2020denoising} to demonstrate the effectiveness of discrete modeling. For a fair comparison, we adopted the same training protocols as in our paper, including the dataset, augmentation methods, batch size, learning rate, optimization methods, and other hyperparameters. We also employed a U-Net~\cite{ronneberger2015u} with a similar architecture to ensure comparable model capacity to our method. The primary difference is that the continuous diffusion model predicts a continuous tensor at each iteration, and the output of the final iteration is binarized using a fixed threshold of 0.5. After training, we evaluated the performance. Since the legality of the generated patterns is guaranteed by the proposed deep squish tensor and legalization method, we focused on comparing the diversity of the generated patterns. As shown in \Cref{tab:abla-discrete}, discrete modeling improves diversity by a reasonable margin (11.294 $\rightarrow$ 11.713). The results support our claim in the main text and indicate that discrete modeling of the topology tensor benefits the layout pattern generation task.

\begin{table}[tb!]
    \centering
    \caption{{Comparison between discrete modeling and continual modeling in topology tensor generation.} }
    \label{tab:abla-discrete}
    {
        \begin{tabular}{c|ccc}
            \toprule
            Modeling & Generated Topology & Diversity \\
            \midrule
            Continual & 100000& 11.294 \\
            Discrete &  100000 &\textbf{11.713}\\
            
            \bottomrule 
        \end{tabular}
        }
\end{table}

\minisection{Probability of Concatenate and Crop Augmentation}
We conduct an ablation study to investigate the impact of the probability of concatenate and crop augmentation on the final results. By gradually increasing the augmentation probability from 0.0 to 1.0 during training, we retrain the model on the augmented data and evaluate the diversity of the generated samples. The results, presented in \Cref{fig:abla-pro}, indicate that the optimal probability range lies within $[0.5,0.9]$. Based on these findings, we set the probability to 0.5 in our implementation.

\minisection{Fast-Sampling Factor \( m \).} In our method, the fast-sampling factor \( m \) can be adjusted to balance efficiency and model performance. We extend the discussion in the main text by testing various values of \( m \). The results are presented in \Cref{tab:efficiency-sample}. We observe that \( m = 10 \) provides a good trade-off between efficiency and model performance. For cases where \( m > 10 \), the diversity of generated samples decreases significantly.

}

\subsection{Discussion on Validity}

A metric referred to as pattern validity was introduced in previous work \cite{zhang2020layout}. This metric is evaluated using an encoder-decoder model pre-trained on the training data. The underlying assumption is that generated patterns that resemble those in the training set will achieve higher scores in this validity metric. However, we contend that this interpretation of validity is somewhat narrow. One of the key objectives in layout pattern generation is to produce a diverse array of legal patterns for various downstream tasks, such as hotspot detection or lithography simulation. In these cases, legal patterns that diverge from the training set are often more desirable, yet they tend to receive lower scores under the current validity metric.

What further undermines the usefulness of this metric is that it promotes overfitting to the training set. For example, as noted in \cite{zhang2020layout,wen2022layoutransformer}, the generated patterns can obtain significantly higher validity scores (from 65\% to 84\%) compared to patterns in the test set, which should share the same distribution as the training data. It is unrealistic to assume that a higher validity score necessarily correlates with better-quality patterns.

For these reasons, we have opted not to assess \tool{DiffPattern}-Flex using this metric.

\section{Conclusion}
\label{sec:conclusion}

In this paper, we proposed \tool{DiffPattern}-Flex, a novel framework for efficient and legal layout pattern generation. By decoupling topology generation and pattern legalization, our method provides flexibility in handling changing design rules without retraining the model, making it highly adaptable for various downstream tasks in design automation. Our approach demonstrates significant improvements in both the diversity and legality of generated patterns, achieving state-of-the-art performance. The ability to generate diverse legal patterns enhances the robustness of machine-learning workflows and supports a wide range of applications, such as lithography simulation and hotspot detection. Future work will focus on expanding \tool{DiffPattern}-Flex to more complex tasks, such as multi-source pattern generation and integration of additional design constraints.

{
    \bibliographystyle{IEEEtran}
    \bibliography{ref/Top,ref/FPGA-DNN,ref/FPGA,ref/FPGAPlacement,ref/tools,ref/DiffPattern}
}

\vspace{-.1in}
\begin{IEEEbiography}[{\includegraphics[height=1.26in,clip,keepaspectratio]{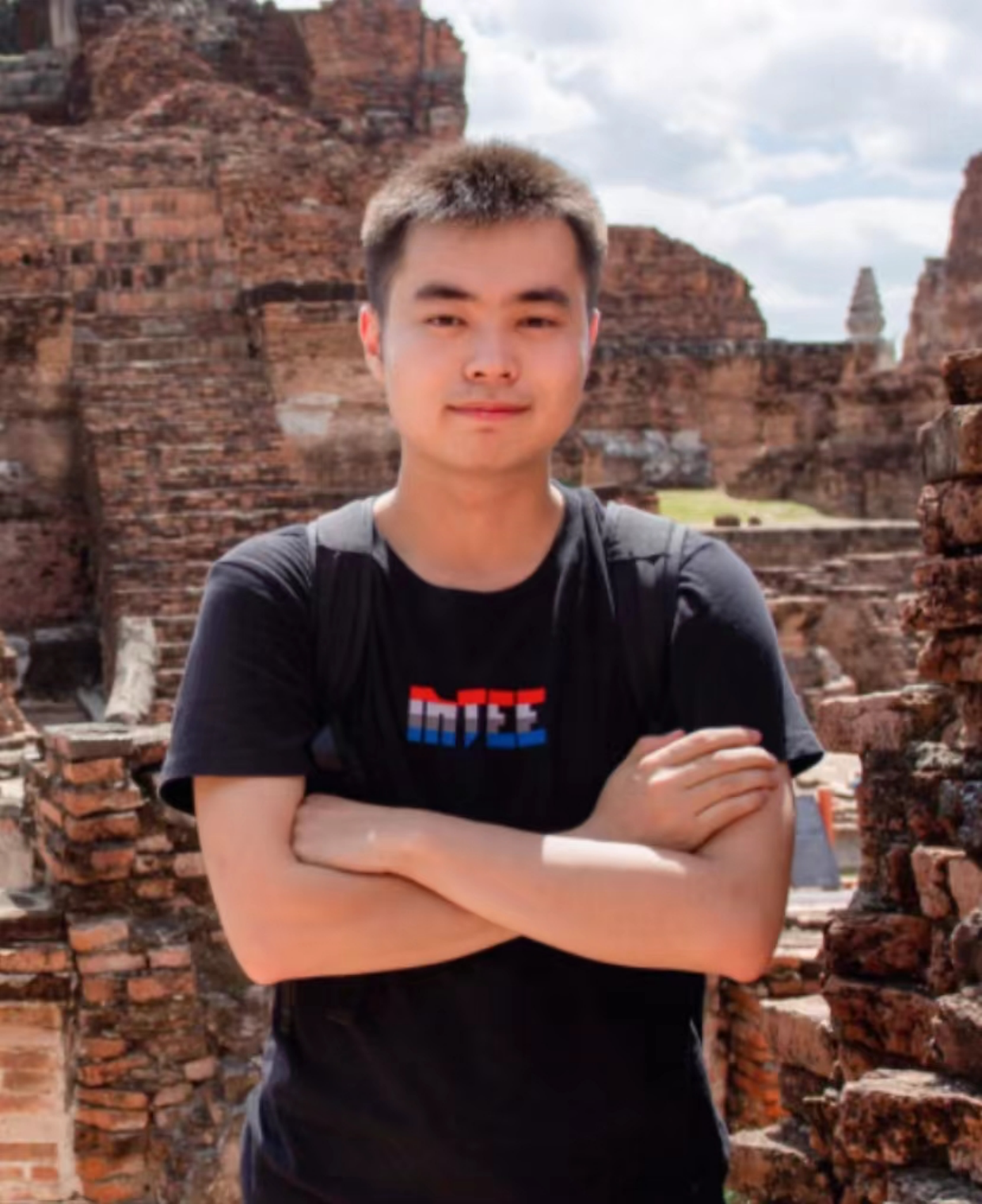}}]
  {Zixiao Wang} received his B.Eng. degree in Automation from Tsinghua University (THU) in 2019 and an MSc.degree in Computer Science from Tsinghua University (THU) in 2022. He is currently pursuing his Ph.D. degree at the Department of Computer Science and Engineering, The Chinese University of Hong Kong. His research interests include Generative AI empowers EDA.
\end{IEEEbiography}

\vspace{-.1in}
\begin{IEEEbiography}[{\includegraphics[height=1.26in,clip,keepaspectratio]{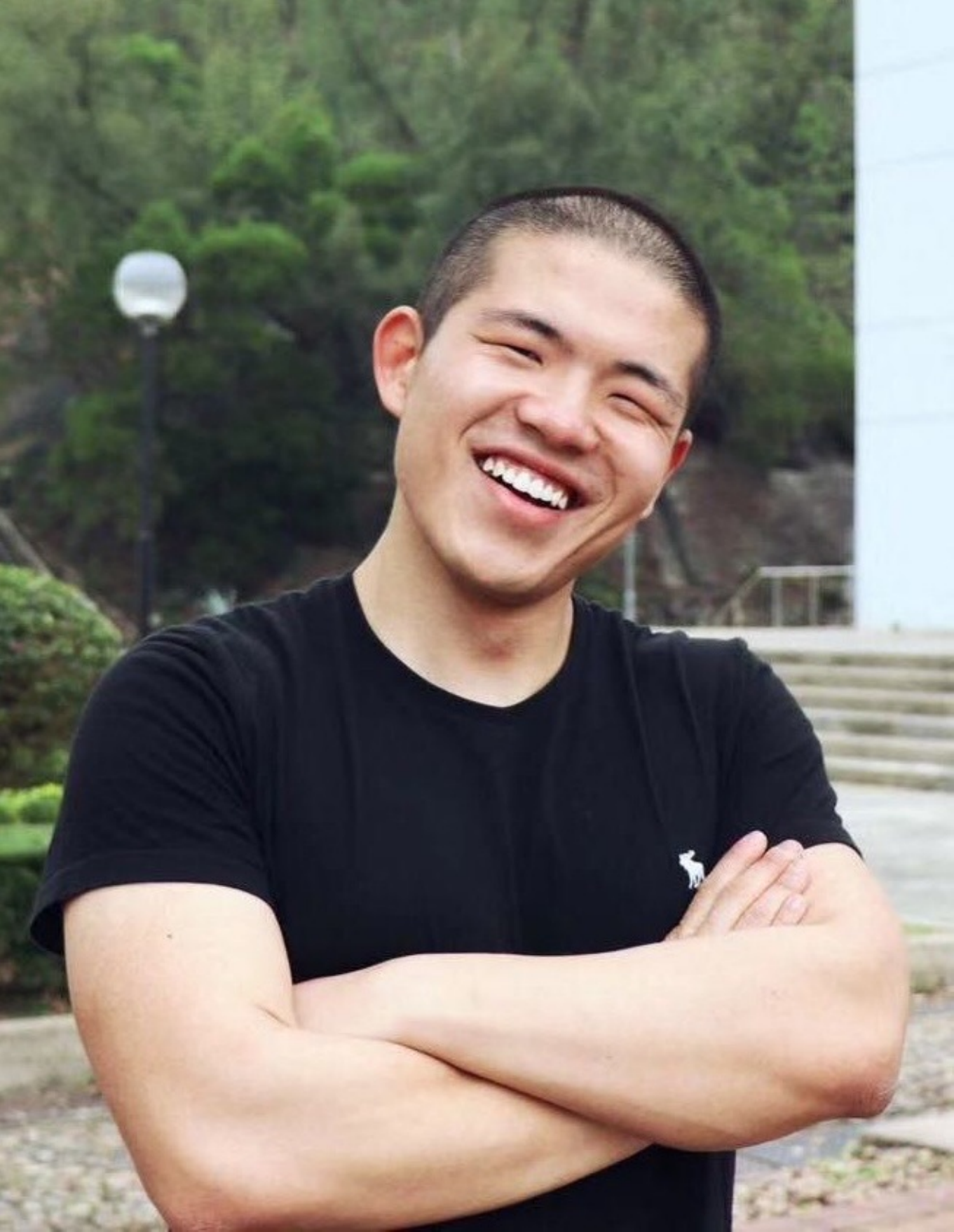}}]
{Wenqian Zhao} obtained his  Ph.D. and B.Sc. of Computer Science and Engineering from The Chinese University of Hong Kong, Hong Kong, in 2024 and 2019. His research interests include machine learning for VLSI design automation and hardware-aware deep- learning acceleration.
\end{IEEEbiography}

\vspace{-.1in}
\begin{IEEEbiography}[{\includegraphics[height=1.26in,clip,keepaspectratio]{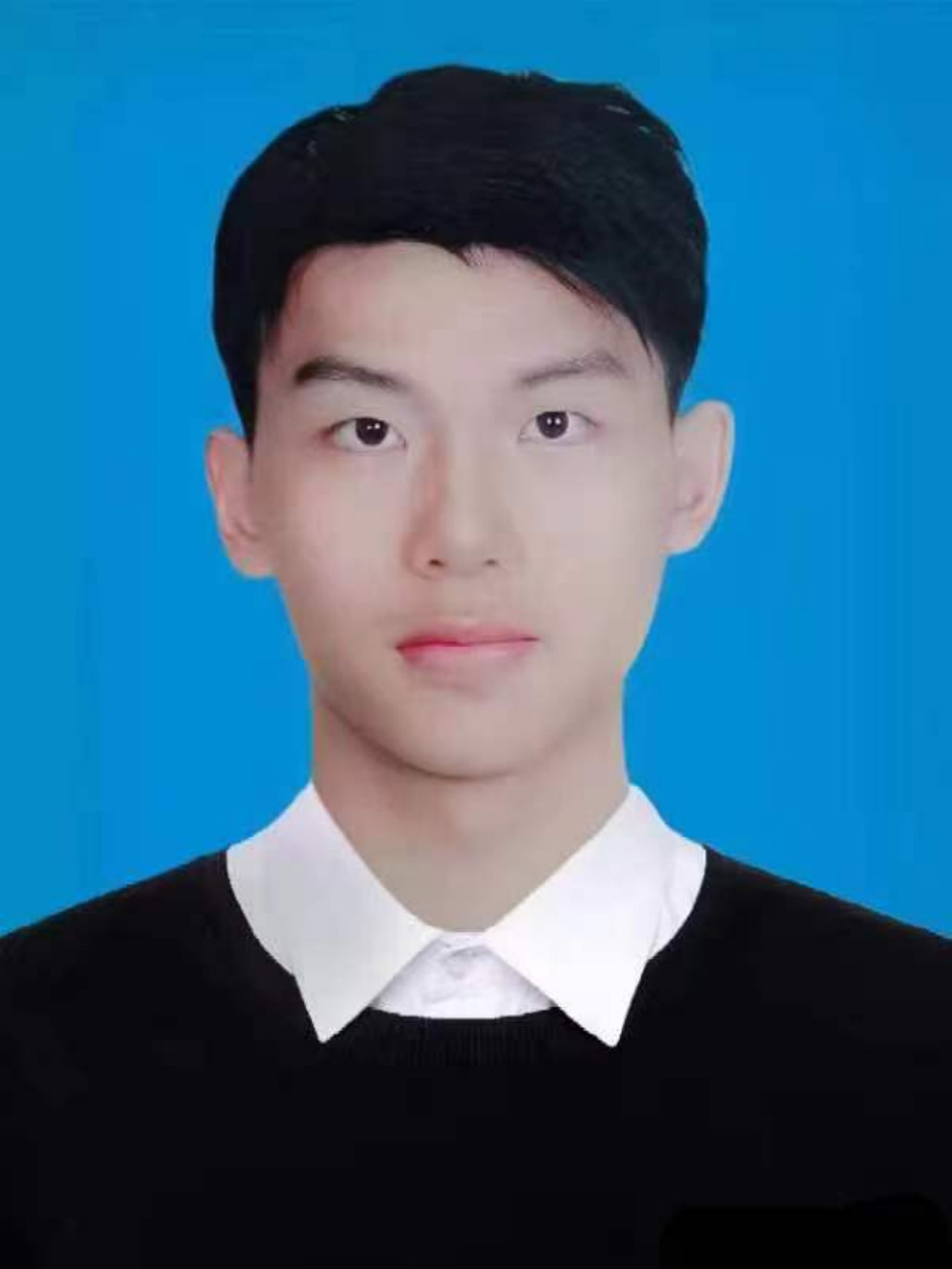}}]
  {Yunheng Shen} received his B.Eng. degree in Automation from Tsinghua University (THU) in 2019 and is currently pursuing his Ph.D. degree at the Department of Automation, THU. His current research interests include federated learning and the application of generative models or large models at the edge.
\end{IEEEbiography}

\vspace{-.1in}
\begin{IEEEbiography}[{\includegraphics[height=1.26in,clip,keepaspectratio]{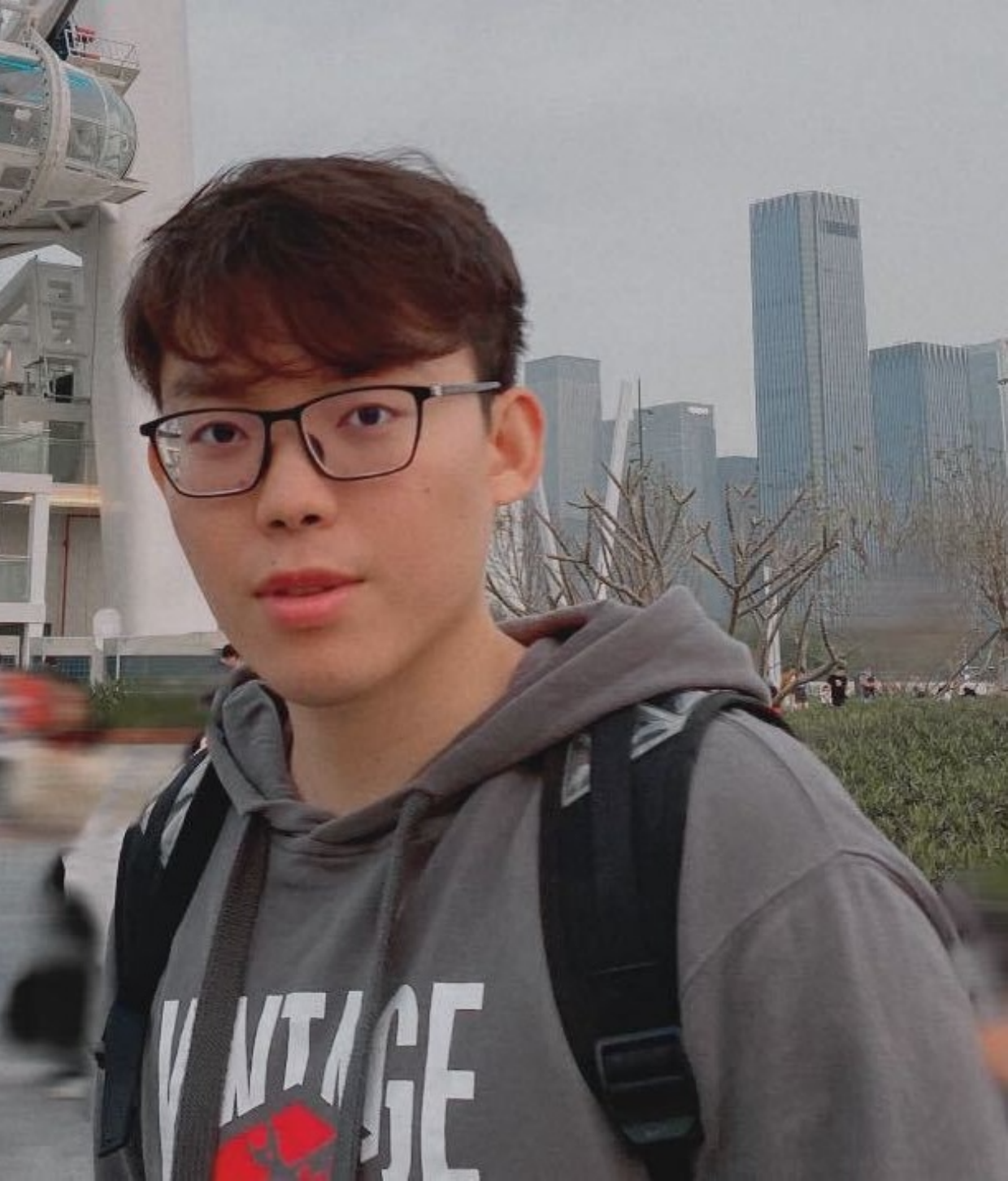}}]
{Yang Bai} received
  the BS degree in telecommunications engineering from Xidian University in 2017, and Master degree in computer science from the Chinese Academy of Sciences
  University in 2020. He is a Ph.D. in Department of Computer
  Science and Engineering, the Chinese University of Hong Kong. His research interests focus on the optimization for deep neural network training and inference via compilation techniques.
\end{IEEEbiography}

\vspace{-.1in}
\begin{IEEEbiography}[{\includegraphics[height=1.26in,clip,keepaspectratio]{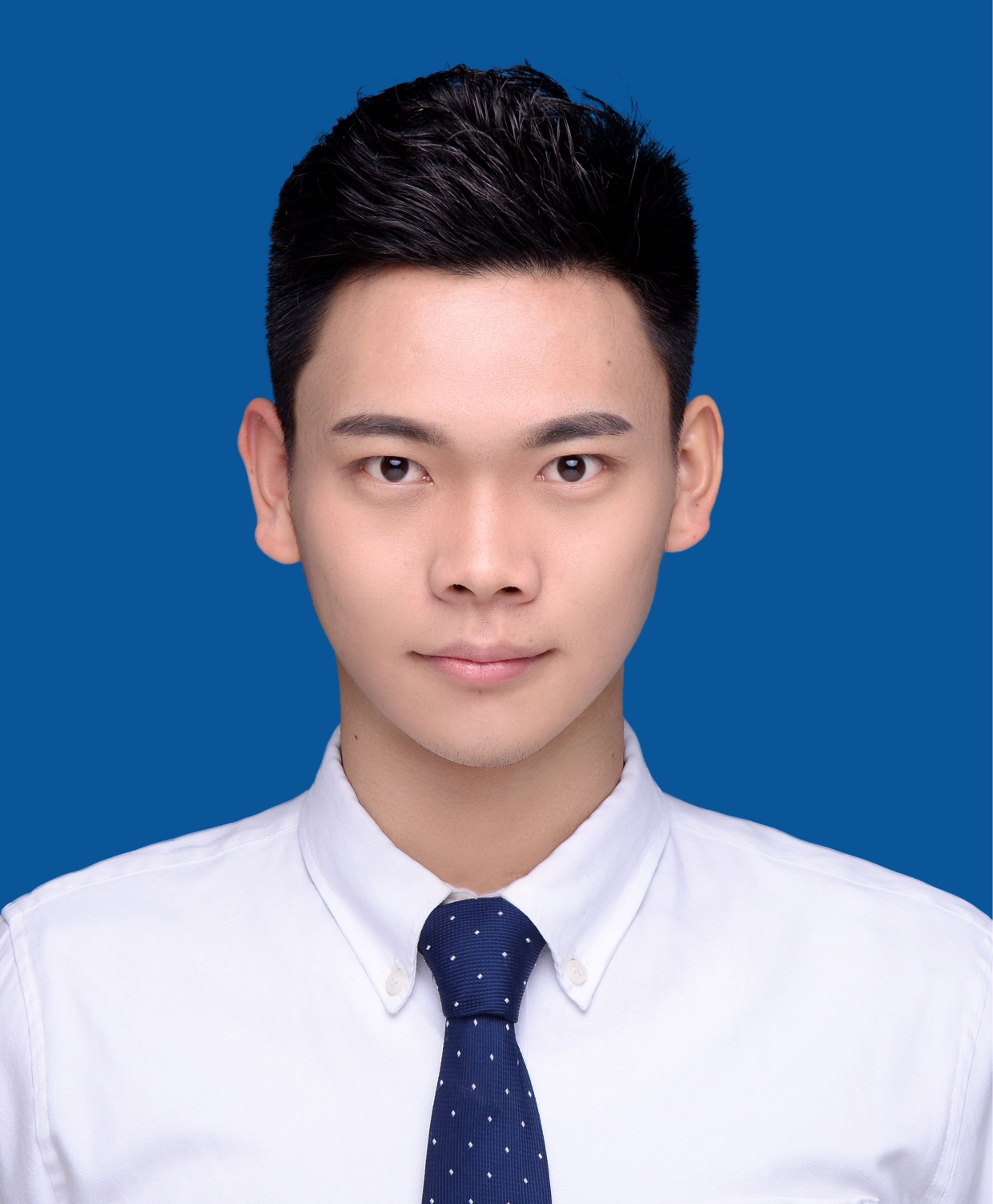}}]
    {Guojin Chen}
    received his B.Eng.~degree in software engineering from Huazhong University of Science and Technology (HUST) in 2019.
    He is currently pursuing his Ph.D. degree at the Department of Computer Science and Engineering, The Chinese University of Hong Kong.
    His current research interests include (1) machine learning in VLSI design for manufacturability and (2) physics-informed networks for solving EDA area problems.
\end{IEEEbiography}

\vspace{-.1in}
\begin{IEEEbiography}[{\includegraphics[height=1.26in,clip,keepaspectratio]{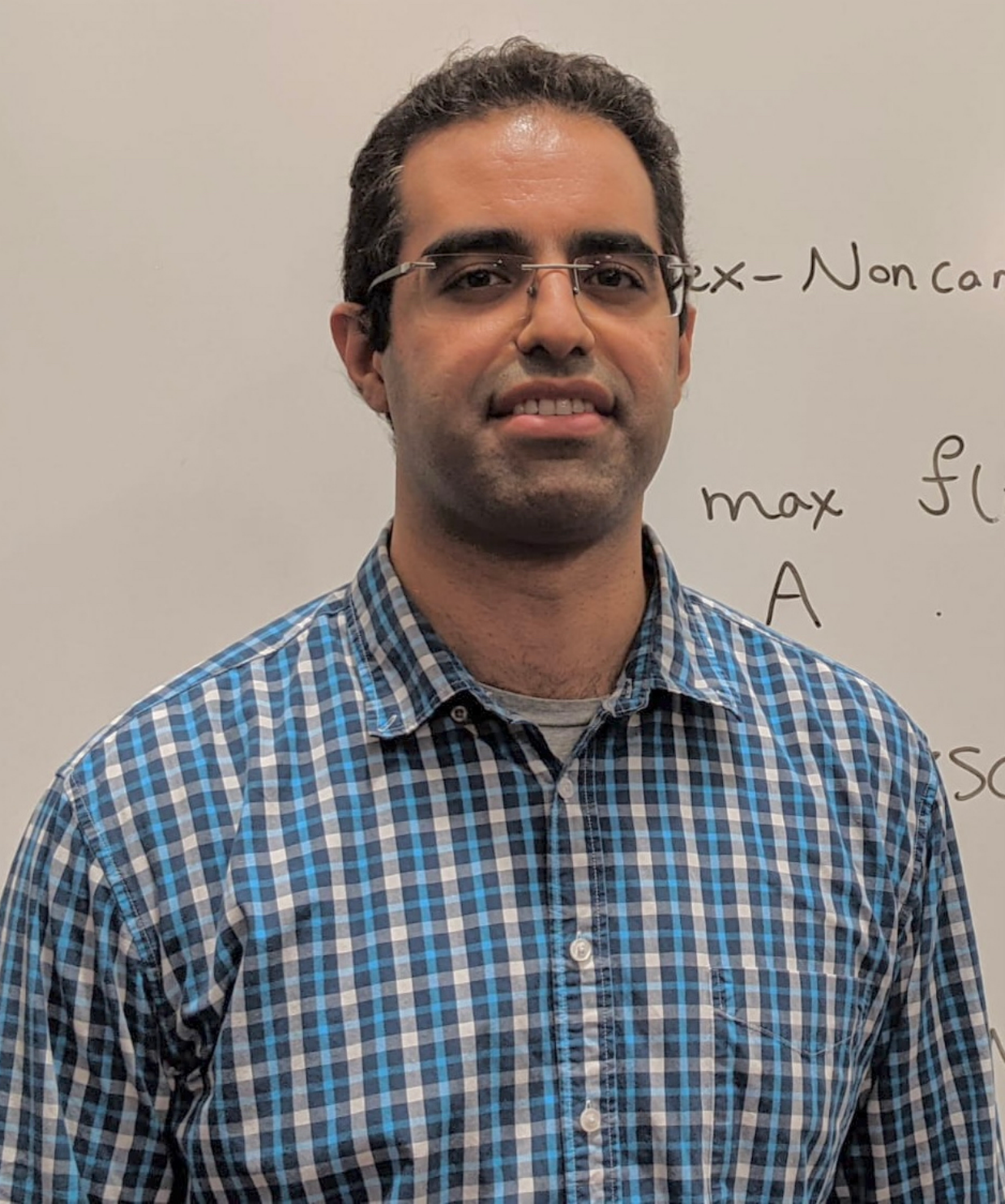}}]
    {Farzan Farnia}
    is an Assistant Professor of Computer Science and Engineering at The Chinese University of Hong Kong. Prior to joining CUHK, he was a postdoctoral research associate at the Laboratory for Information \& Decision Systems, Massachusetts Institute of Technology, from 2019-2021. He received his M.Sc. and Ph.D. degrees in Electrical Engineering from Stanford University where he was a graduate research assistant at the Information Systems Laboratory advised by David Tse. He also received his B.Sc. degree in Electrical Engineering and Mathematics from Sharif University of Technology. His research interests lie in learning and information sciences with a particular focus on multi-learner learning frameworks where he study the convergence, equilibrium, and robustness properties of multi-learner learning algorithms. 
\end{IEEEbiography}

\vspace{-.1in}
\begin{IEEEbiography}[{\includegraphics[height=1.26in,clip,keepaspectratio]{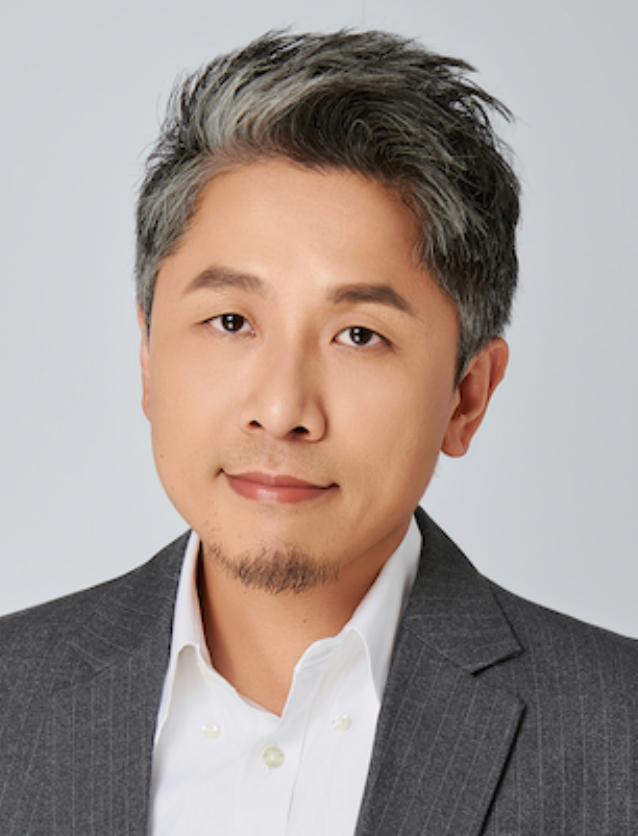}}]
    {Bei Yu}
    (M'15-SM'22)
    received the Ph.D.~degree from The University of Texas at Austin in 2014.
    He is currently an Associate Professor in the Department of Computer Science and Engineering, The Chinese University of Hong Kong.
    He has served as TPC Chair of ACM/IEEE Workshop on Machine Learning for CAD, and in many journal editorial boards and conference committees.
    He received eleven Best Paper Awards from ICCAD 2024 \& 2021 \& 2013,
    IEEE TSM 2022, DATE 2022, ASPDAC 2021 \& 2012, ICTAI 2019, Integration, the VLSI Journal in 2018,
    ISPD 2017, SPIE Advanced Lithography Conference 2016, six ICCAD/ISPD contest awards,
    IEEE CEDA Ernest S. Kuh Early Career Award in 2021,
    DAC Under-40 Innovator Award in 2024, 
    and Hong Kong RGC Research Fellowship Scheme (RFS) Award in 2024.
\end{IEEEbiography}

\end{document}